\documentclass{article}

\usepackage{microtype}
\usepackage{graphicx}
\usepackage{subfigure}
\usepackage{booktabs} 

\usepackage{hyperref}

\usepackage[accepted]{icml2024}

\usepackage{amsmath}
\usepackage{amssymb}
\usepackage{mathtools}
\usepackage{amsthm}
\usepackage{bm}
\usepackage{bbm}
\usepackage{multirow}
\usepackage{xcolor}

\usepackage[capitalize,noabbrev]{cleveref}

\theoremstyle{plain}
\newtheorem{theorem}{Theorem}[section]
\newtheorem{proposition}[theorem]{Proposition}
\newtheorem{lemma}[theorem]{Lemma}

\theoremstyle{definition}

\newtheorem{assumption}[theorem]{Assumption}
\theoremstyle{remark}

\usepackage[textsize=tiny]{todonotes}

\icmltitlerunning{Adaptive Advantage-Guided Policy Regularization for Offline Reinforcement Learning}

\begin{document}

\twocolumn[
\icmltitle{Adaptive Advantage-Guided Policy Regularization \\for Offline Reinforcement Learning}



\icmlsetsymbol{equal}{*}

\begin{icmlauthorlist}
\icmlauthor{Tenglong Liu}{yyy}
\icmlauthor{Yang Li}{comp}
\icmlauthor{Yixing Lan}{yyy}
\icmlauthor{Hao Gao}{yyy}
\icmlauthor{Wei Pan}{comp1}
\icmlauthor{Xin Xu}{yyy}
\end{icmlauthorlist}

\icmlaffiliation{yyy}{National University of Defense Technology, Changsha, China}
\icmlaffiliation{comp}{Shanghai Institute of Microsystem and Information Technology, Chinese Academy of Sciences, Shanghai, China}
\icmlaffiliation{comp1}{Delft University of Technology, The Netherlands}

\icmlcorrespondingauthor{Xin Xu}{Xinxu@nudt.edu.cn}
\icmlcorrespondingauthor{Yixing Lan}{lanyixing16@nudt.edu.cn}

\icmlkeywords{Machine Learning, ICML}

\vskip 0.3in
]



\printAffiliationsAndNotice{}  

\begin{abstract}
In offline reinforcement learning,  the challenge of out-of-distribution (OOD) is pronounced. To address this, existing methods often constrain the learned policy through policy regularization. However, these methods often suffer from the issue of unnecessary conservativeness, hampering policy improvement. This occurs due to the indiscriminate use of all actions from the behavior policy that generates the offline dataset as constraints.
The problem becomes particularly noticeable when the quality of the dataset is suboptimal. Thus, we propose Adaptive Advantage-Guided Policy Regularization (A2PR), obtaining high-advantage actions from an augmented behavior policy combined with VAE to guide the learned policy.  A2PR can select high-advantage actions that differ from those present in the dataset, while still effectively maintaining conservatism from OOD actions.
This is achieved by harnessing the VAE capacity to generate samples matching the distribution of the data points. We theoretically prove that the improvement of the behavior policy is guaranteed. Besides, it effectively mitigates value overestimation with a bounded performance gap. Empirically, we conduct a series of experiments on the D4RL benchmark, where A2PR demonstrates state-of-the-art performance. Furthermore, experimental results on additional suboptimal mixed datasets reveal that A2PR exhibits superior performance. Code is available at \url{https://github.com/ltlhuuu/A2PR}.
\end{abstract}

\section{Introduction}
\label{introduction}
    Reinforcement learning 
    has made substantial breakthrough advancements over the past decades, such as chess games~\cite{silver2016mastering, schrittwieser2020mastering, li2023jiangjun}, video games~\cite{perolat2022mastering, AlphaStar}, robotics~\cite{hwangbo2019learning,andrychowicz2020learning,rajeswaran2017learning} and so on.
    Specifically, RL employs a trial-and-error approach, iteratively refining its performance through interactions with the environment in an online fashion~\cite{sutton2018reinforcement}.
    However, the trial-and-error paradigm poses significant challenges to the seamless integration of RL into real-world applications such as autonomous driving, healthcare, and other tasks. The impracticality of trial-and-error in scenarios where active interaction with the environment is unfeasible renders each training run unrealistic. In recent times, offline RL has garnered considerable attention for its potential to exclusively learn from pre-collected datasets, eliminating the need for real-time interaction during training~\cite{levine2020offline}.
    
    In offline RL, a key challenge is addressing the overestimation of Q-values caused by out-of-distribution (OOD) actions. Commonly, techniques rely on incorporating the dataset's behavior policy to tackle this challenge, constraining the learned policy through policy regularization methods~\cite{levine2020offline}. These regularization methods introduce an extra term to calculate divergence metrics between the learned policy and the behavior policy, employing widely-used metrics such as behavior clone~\cite{fujimoto2021minimalist, ran2023policy}, Kullback-Leibler (KL) divergence~\cite{jaques2019way, wu2019behavior}, fisher divergence~\cite{kostrikov2021offline1}, and Maximum Mean Discrepancy (MMD)~\cite{kumar2019stabilizing}. To some extent, these methods alleviate overestimation from OOD actions~\cite{levine2020offline}. However, existing policy regularization methods are unnecessarily conservative~\cite{hong2023beyond} since they force the learned policy to closely mimic actions of the behavior policy, even if those actions are suboptimal. Such unnecessary conservatism hampers policy performance, especially in datasets dominated by low-return trajectories with sparse high-return instances.
    
    To address the issue of unnecessary conservatism~\cite{hong2023beyond}, we introduce an Adaptive Advantage-Guided Policy Regularization (A2PR) method for offline RL. This approach combines policy regularization with high-advantage actions and efficiently guides the learned policy toward improvement. In A2PR, a Variational Autoencoder (VAE) enhanced by an advantage function generates high-advantage actions by combining them with those from the dataset for policy regularization. This process is similar to directing the learned policy with sensible behavior from an augmented behavior policy. A2PR has the flexibility to select high-advantage actions differing from those in the dataset due to the augmented behavior policy. Additionally, it inherently exhibits a natural level of conservatism since the VAE generates samples from the same distribution as the data points. In contrast to prior methods that force the learned policy to closely mimic all data, A2PR promotes policy improvement while being guided by superior actions from the augmented behavior policy. It eases the constraint when encountering poorer data, alleviating the pessimistic issues associated with overly conservative constraints. Consequently, A2PR establishes a more effective and adaptive policy constraint.

    A2PR can integrate into existing actor-critic offline RL algorithms. In our study, we implement a practical algorithm based on TD3~\cite{fujimoto2018addressing}, chosen for its straightforward and efficient implementation that yields remarkable performance. 
    Subsequently, we perform a theoretical analysis investigating the performance improvement of the behavior policy. Our findings illustrate a reduction in the overestimation problem, substantiated by quantifying a bounded performance gap concerning the learned policy.
    The experimental results show that our proposed method attains state-of-the-art performance on the D4RL standard benchmark~\cite{fu2020d4rl} for offline RL. Furthermore, we assess the method's performance on supplementary low-quality datasets, comprised of 99\% random policy datasets. Our approach exhibits noteworthy performance improvements, particularly evident in the additional suboptimal or low-quality datasets.
\section{Preliminaries}
\label{sec2}

    This section provides a concise introduction to the background and introduces some key notation. Offline RL, also referred to as batch RL or data-driven RL~\cite{levine2020offline}, constitutes a specialized category within RL. It operates within the framework of Markov decision processes (MDPs) denoted as $(\mathcal{S, A}, P, r, \gamma)$~\cite{sutton2018reinforcement}, where $\mathcal{S}$ represents the state space, $\mathcal{A}$ denotes the action space, $P(\cdot|s,a)$ characterizes the transition probability distribution function, $\gamma$ is the discount factor, and $r(s,a)$ corresponds to the reward function for $(s,a)$. Throughout, we consider $\forall (s,a)\in \mathcal{S \times A}, \gamma\in (0,1],|r(s,a)|\leq R_{max}$, and $a\in [-A,A]$. The objective is to identify a policy $\pi^*$ that maximizes the expected return, commencing from any state $s\in S$: $\pi^* = \arg\max_\pi \mathbb E_\pi\left[\sum_{t=0}^\infty\gamma^t r(s_t,a_t)\right]$.

    Offline RL involves learning policies from a predetermined dataset $\mathcal{D}\{(s,a,r,s')\}$ gathered in advance through an unknown behavior policy $\pi_\beta$. This approach enhances sample efficiency by leveraging pre-collected data without requiring extensive direct interaction with the environment. Offline RL holds notable significance, especially in situations where interaction entails risks or incurs high costs. Consequently, the difference between the learned policy and the behavior policy often gives rise to  OOD actions, leading to extrapolation error.

    The Q-function $Q(s,a)$ signifies the expected discounted return starting from any state $s\in S$. The advantage function of the action $a$ is defined as: $A(s,a)=Q(s,a) - V(s)$, where $V(s)$ represents the value function. For each policy $\pi$, there exists a corresponding Q-function obtained through the Bellman operator $\mathcal{T}$, defined as:
    \begin{equation}
    \vspace{-1mm}
    \begin{split}
        Q^\pi(s,a) &= \mathbb E_\pi\left[\sum_{t=0}^\infty\gamma^t r(s_t,a_t)|s_0=s,a_0=a\right], \\
        (\mathcal{T} Q^\pi)(s,a) &= r(s,a) + \gamma \mathbb E_{\substack{s' \sim P(\cdot|s,a),\\
        a' \sim\pi^*(\cdot|s')}}[Q(s',a')].
    \end{split}
    \vspace{-1mm}
    \end{equation}
    Subsequently, the expected discounted reward $J(\pi)$ of policy $\pi$ can be expressed as $J(\pi) = \frac{1}{1-\gamma}\mathbb E_{s\sim d_\pi(s)}[r(s)]$. The discounted state distribution of a policy $\pi$, also known as the occupancy measure $d_\pi$: $\mathcal{S}\rightarrow{\mathbb R}$, is defined as $d_\pi(s) = \sum_{t=0}^\infty\gamma^t p(s_t=s|\pi)$, where $p(s_t=s|\pi)$ represents the probability of state $s_t$ being $s$ under policy $\pi$.

    The objective values utilized in Bellman backups for policy evaluation originate from actions sampled from the learned policy $\pi$. However, the Q-function is exclusively trained using actions sampled from the behavior policy $\pi_\beta$, responsible for generating the dataset $\mathcal{D}$. As $\pi$ is optimized to maximize Q-values, a potential bias towards OOD actions may exist, resulting in inaccurately overestimated Q-values. We define $\delta_{error}$ as the overestimation error~\cite{fujimoto2019off}, representing the disparity between an approximate estimate $\tilde{Q}^\pi$ and the true Q-value function $Q^\pi$: $\delta_{error} = \tilde{Q}^\pi(s,a) - Q^\pi(s,a)$.

\section{Related Work}
\label{sec_relat}
\subsection{Offline RL with policy regularization.}
    Policy regularization is pivotal research in offline RL, addressing distribution shift challenges to mitigate OOD actions. TD3+BC~\cite{fujimoto2021minimalist} enhances policy improvement by integrating a straightforward behavior cloning term, providing a clear estimate of the learned policy.
    Numerous divergence penalties compel the learned policy to stay close to the behavior policy, including Maximum Mean Discrepancy (MMD)~\cite{kumar2019stabilizing}, Fisher divergence~\cite{kostrikov2021offline1}, KL divergence~\cite{jaques2019way, wu2019behavior, nair2020awac}, and Wasserstein distance~\cite{wu2019behavior}.
    BEAR~\cite{kumar2019stabilizing} employs MMD with a Gaussian kernel as divergence regularization for policy improvement but relies heavily on the approximate nature of low-sampled MMD. Advantage-weighted~\cite{peng2019advantage} regression utilizes supervised regression as learning subroutines to enhance the learned policy while concurrently enforcing an implicit KL-divergence constraint.
    BCQ~\cite{fujimoto2019off} and LAPO~\cite{chen2022latent} implement the modeling of the behavior policy using Conditional VAE~\cite{sohn2015learning}, achieving proximity between the learned and behavior policies through an implicit constraint.
    SPOT~\cite{wu2022supported} explicitly pre-trains a VAE to model the support set of the behavior policy, directly using behavior density to constrain the learned policy.
    OAP~\cite{yang2023boosting} utilizes complex RankNet pseudo-queries to select actions with higher Q-values for policy constraints from the current policy and dataset.
    PRDC~\cite{ran2023policy} employs KD-tree~\cite{bentley1975multidimensional} to index potential actions corresponding to the current state from the entire dataset. However, it's worth noting that the use of KD-tree introduces a high computational complexity. Unlike PRDC, our methods use a simple and effective VAE to generate more potential high-advantage actions for policy regularization.

\subsection{Data reweighting}
    In offline RL, ReDs~\cite{singh2023reds} focuses on reweighting the data distribution solely for CQL~\cite{kumar2020conservative}, aiming to achieve an approximate support constraint formulation. RB-CQL~\cite{jiang2023offline} specializes in the context of CQL by incorporating a retrieval process that recalls past related experiences. AW~\cite{hong2023harnessing} reweights trajectories based on their returns, a process that necessitates the consideration of entire trajectories. DW~\cite{hong2023beyond} emulates sampling from an alternate dataset through importance sampling, where the weighting function can be interpreted as the density ratio between the alternative dataset and the original one.
    OPER~\cite{yue2023offline} adopts a priority function to prioritize a dataset for the enhancement of the learned policy.
    In the realm of offline imitating learning (IL)~\cite{kim2021demodice,ma2022versatile,xu2022discriminator}, the focus is on training an expert policy from a dataset that comprises a limited set of expert data along with a substantial amount of random data. These methods aim to train a model to learn a policy that closely aligns with the expert data, necessitating the separation of expert data from random data. In contrast, our methods solely require the advantage of selecting actions without the need for separable data.

\section{Method}
\label{sec_method}
    In this section, we present the A2PR algorithm. We commence by introducing elevating positive behavior learning in Section~\ref{EPBL}, aiming to procure more high-advantage actions with the support of the dataset. Subsequently, in Section~\ref{AAPC}, we delve into the adaptive advantage policy constraint, designed to reinforce policy improvement by prioritizing actions with higher advantages. The practical implementation details of the algorithm are outlined in Section~\ref{prac_imple}. Finally, in Section~\ref{theoretical1}, we provide theoretical analyses elucidating the performance improvement guarantee over the behavior policy and a bounded performance gap, addressing the overestimation issue.

\subsection{Elevating Positive Behavior Learning}\label{EPBL}

    In offline RL, the sampled probability of actions remains fixed due to the unchanging nature of the pre-collected dataset. Notably, existing offline RL methods~\cite{hong2023harnessing} demonstrate that reweighting datasets based on trajectory return or episode advantage effectively regulates the implicit behavior policy, resulting in enhanced performance. This approach serves to provide a more favorable starting performance for the learned policy. Motivated by these findings, it is intuitive to consider that augmenting the probability of high-advantage actions can contribute to the improvement of the implicit behavior policy. Furthermore, we delve into a theoretical analysis guaranteeing the enhancement of the behavior policy.
    \begin{proposition}
    \label{proposition1}
    Suppose that $A^{\pi_\beta}(s,a)(\hat{\pi}_\beta(a|s)-\pi_\beta(a|s)) \geq 0$. Then, we have
    \begin{equation}
    \vspace{-1.5mm}
    \begin{aligned}
        J(\hat{\pi}_\beta) - J(\pi_\beta) \geq 0 ,\\ 
    \end{aligned}
    \vspace{-1mm}
    \end{equation}
    \end{proposition}
    where $\hat{\pi}_\beta$ is another behavior policy, $\pi_\beta$ is the original behavior policy of the dataset. The proof is deferred to Appendix~\ref{appendix:A1}.
    
    Using a VAE as the density estimator for the dataset~\cite{kumar2019stabilizing, wu2022supported} proves to be a straightforward and effective method for learning the behavior policy. In this regard, we propose a specific approach to optimize the VAE output action in conjunction with the advantage function. In the realm of offline RL, where $A(s,a)$ represents the additional reward achievable by taking action $a$ rather than the expected return, it serves as a quality indicator for actions. This information can be effectively incorporated with the reconstruction component of the VAE. Motivated by this insight, we introduce elevating positive behavior learning (EPBL) utilizing a VAE enhanced by the advantage function.
    The EPBL can be optimized jointly with the following evidence lower bound (ELBO):
    \begin{equation}\label{elbo1}
    \begin{split}
       \log p_{\psi}(a|s) &\geq \mathbb{E}_{q_{\varphi}(z|a,s)} \left[
        f(A(s,a)>\epsilon_A) \log p_{\psi}(a|z,s) \right] \\
        &\qquad\qquad\qquad\quad- \text{KL} \left[ q_{\varphi}(z|a,s) \parallel p(z|s) \right],
    \end{split}
    \end{equation}
    where $f(A(s,a)>\epsilon_A)$ represents $w_1 * {\mathbbm 1}(A(s,a)>\epsilon_A)$, $w_1$ is a hyperparameter.
    $A(s,a)$ represents the advantage of action $a$, $\log p_{\psi}(a|z,s)$ signifies the likelihood measure of the reconstructed action from the decoder, and $\text{KL} \left[ q_{\varphi}(z|a,s) \parallel p(z|s) \right]$ represents the KL-divergence between the encoder output and the prior of $z$. $\epsilon_A$ is an advantage threshold. This threshold serves to constrain the quality of the reconstructed action.
    Restricting the reconstruction process to actions with a higher advantage from the dataset is achieved by introducing a simple constraint $ {\mathbbm 1} (A(s,a)>\epsilon_A)$ before the reconstruction loss. This constraint ensures the reconstruction of actions with a higher advantage. As a result, the enhanced behavior policy $\hat\pi_\beta$ assigns a greater density ratio to high-advantage actions compared to the behavior policy $\pi_\beta$.

\subsection{Adaptive Advantage Policy Constraint}\label{AAPC}

    A2PR faces the challenge of reconciling two conflicting objectives~\cite{yang2023boosting}: policy improvement and policy constraint. An overly strict policy constraint within a suboptimal dataset may impede the policy's enhancement beyond the behavior policy. Conversely, a lax constraint might lead to distributional shift, causing the learned policy to fail in OOD actions. Achieving a balance between these aspects is imperative. By incorporating a policy constraint that prioritizes actions with a higher advantage, alongside policy improvement considerations, A2PR enables the learned policy to assimilate knowledge from the augmented behavior policy. These dual approaches of policy improvement and policy constraint can be expressed generically through the following equation:
        \begin{equation}
        \begin{split}\label{eq1}
         \max_{\phi}\mathbb E_{s \sim \mathcal{D}}[Q_{\theta}(s,\pi_{\phi}(s))] \quad &\Longrightarrow \text{Policy improvement},\\
              \mathrm{ s.t.}\quad ||\pi_{\phi}(s)-a|| < \epsilon_0 \qquad &\Longrightarrow\text{Policy constraint},\\
        \end{split}
        \end{equation}
    where $Q_\theta$ represents the state-action value function, $\pi_\phi(\cdot)$ denotes the learned policy, and $||\cdot||$ stands for a norm.
    Selecting high-advantage actions from the augmented behavior policy is pivotal for policy regularization:
    \begin{equation}
    \begin{split}\label{eq0}
        \tilde{a}&=\mathop{\arg\max}_{\dot a\in\{a,\hat\pi_\beta(s)\}}A(s,\dot a),\\
    \end{split}
    \end{equation}
    where $\tilde{a}$ represents the high-advantage action, chosen between $a$ and the generative high-advantage actions $\hat{\pi}_\beta(s)$.
    
    To be more specific, if the actions adhere to the condition $A(s,\tilde{a}) \geq \epsilon_A$, the policy will be constrained in proximity to $\tilde a$. Conversely, when the condition $A(s,\tilde{a}) \geq \epsilon_A$ is not satisfied, the current policy should self-learn, indicating the necessity for a robust constraint. The advantageous action $\Bar{a}$ is dynamically determined by the following equation:

    \begin{equation}\label{final_a}
    \begin{aligned}
                \Bar{a} &=\left\{
                    \begin{array}{ll}
                      \tilde{a}, \quad\quad A(s,\tilde{a}) \geq \epsilon_A\\
                      \pi_{\phi}(s),A(s,\tilde{a})<\epsilon_A,\\
                    \end{array}
                  \right.\\
    \end{aligned}
    \end{equation}
    
    Once the advantageous action $\Bar{a}$ is determined, the learned policy can adaptively achieve policy constraint, and the objective transforms to:
    \begin{equation}\label{eq_final}
        \mathcal{L}(\phi) = \mathbb E_{\substack{s,a\sim \mathcal{D},\\ \Bar{a}\in\{\tilde{a},\pi_\phi(s)\}}}\left[-\lambda Q_{\theta}(s,\pi_\phi(s)) + 
        (\pi_{\phi}(s)-\Bar{a})^2\right],
    \end{equation}
    where $\lambda$ represents a hyperparameter. A2PR steers the learned policy toward actions with high advantage, fostering improvement. Simultaneously, it dynamically balances the interplay between enhancing policy and imposing constraints. Consequently, A2PR safeguards the learned policy against the influence of suboptimal actions.

\subsection{Practical Implementation}\label{prac_imple}

    Our algorithm framework builds upon TD3+BC. The parameters $\theta_1, \theta_2, \phi, \psi$ pertain to two Q-networks, the policy network, and the value network, respectively. Additionally, $\theta_1^{'}, \theta_2^{'}, \phi^{'}$ correspond to the parameters of the target Q-networks and the target policy network. To achieve a more balanced integration of Q-value and regularization, we formalize the Q-value within the policy loss as follows: $ \mathcal{L}(\phi) = \mathbb E_{\substack{s,a\sim \mathcal{D},\\ \Bar{a}\in\{\tilde{a},\pi_\phi(s)\}}}\left[-\lambda Q_{\theta}(s,\pi_\phi(s)) + (\pi_{\phi}(s)-\Bar{a})^2\right] $. Here, $\lambda = \frac{\alpha N}{\sum_{s_i,a_i}Q(s_i,a_i)}$, where $\alpha$ is a hyperparameter and $N$ represents the batch size~\cite{fujimoto2021minimalist}.
        
    To derive the advantage function, our approach draws inspiration from IQL~\cite{kostrikov2021offline}, which focuses on learning exclusively within the dataset's support to mitigate overestimation issues related to OOD actions. 
    We have similarly customized the learning processes for both the Q-function and the Value-function. Initially, the parameter $\theta$ undergoes optimization by minimizing the following temporal difference (TD) error:
    \begin{equation}\label{eq5}
    \begin{split}
        \mathcal{L}_Q(\theta_i) = \mathbb E_{(s,a,s')\sim \mathcal{D},a'\sim h(s')}[(r(s,a) \\+ \gamma \min_i Q_{\theta_i^{'}}(s',a') - Q_{\theta_i}(s,a))^2], \\
    \end{split}
    \end{equation}
where $Q_{\theta_i^{'}}$ represents a target Q-value function, and $h(s')=\texttt{clip}(\pi_{\phi'}(s')+\epsilon_0, -A, A),\epsilon_0 \sim \texttt{clip}(\mathcal{N}(0,\hat{\sigma}^2), -c, c)^3, i\in{1,2}$. Here, $c$ and $\hat{\sigma}$ denote two hyperparameters for exploration. A distinct value function is employed to approximate an expectile solely concerning the Q-function, leading to the ensuing loss:
    \begin{equation}\label{eq4}
        \mathcal{L}_V(\varepsilon) = \mathbb E_{(s,a)\sim \mathcal{D}}[(Q_{\theta_i}(s,a)-V_\varepsilon(s))^2],
    \end{equation}
    where $V_\varepsilon$ represents the value function. 
    This design ensures the avoidance of excessive conservatism. Subsequently, the equation for the advantage function is derived as:
    \begin{equation}\label{eq6}
        A(s,a) = Q_{\theta_i}(s,a) - V_\varepsilon(s).
    \end{equation}

    Consequently, the policy can take superior actions in states that go beyond the limitations of the dataset, alleviating undue pessimism associated with suboptimal or low-return behaviors arising from unnecessary policy constraints.
    
\begin{algorithm}[tb]
    \caption{Adaptive Advantage-Guided Policy Regularization~(A2PR)}
    \label{alg:cap}
    \begin{algorithmic}
    \STATE {\bfseries Input:} {Replay buffer $\mathcal{D}$, hyper-parameters $\alpha$, batch size $N$, target network update rate $\tau$ .}

    \STATE Initialize two Q networks with $\theta_1, \theta_2$, policy network with $\phi$ and value function network with $\varepsilon$, target Q and target policy network with $\theta_1^{'} \leftarrow{\theta_1}, \theta_2^{'}\leftarrow{\theta_2}, \phi' \leftarrow{\phi}$, VAE networks with $G_{\psi,\varphi}=\{E_{\varphi},D_{\psi} \}$.

        \FOR{$t=1$ {\bfseries to} $T_1$}
        \STATE Sample mini-batch of transitions $(s,a,r,s')\sim \mathcal{D}$\\
        \textbf{Advantage-guide VAE update:}\\
        \quad $\mu, \sigma=E_{\varphi}(s,a), \hat{a}=D_{\psi}(s,z), z\sim \mathcal{N}(\mu, \sigma)$\\
        \quad Update it by minimizing Equation~(\ref{elbo1})\\
        \textbf{Q-function and value-function update:}\\
        \quad Update Q-value by minimizing Equation~(\ref{eq5})\\
        \quad Update Value function by minimizing Equation~(\ref{eq4})\\
        \textbf{Adaptive Advantage Policy update:} \\
        \quad Update policy network by minimizing Equation~(\ref{eq_final})\\
        \textbf{Update Target Networks: }\\
        \qquad $\theta^{'}_i \leftarrow{\tau\theta + (1 - \tau)\theta^{'}_i},i=1,2$
        \ENDFOR
    \end{algorithmic}
\end{algorithm}

\subsection{Theoretical Analysis}\label{theoretical1}

    We provide theoretical validation for the effectiveness of A2PR. Proposition~\ref{proposition2} indicates that the high-advantage action, chosen by the maximum advantage function from the augmented behavior policy, allows for a superior behavior policy to guide the learned policy towards improvement. With adaptive advantage policy regularization, Theorem~\ref{theorem1} illustrates that A2PR can alleviate the Q-value overestimation problem arising from OOD actions. Additionally, Theorem~\ref{theorem2} highlights a performance gap between the optimal policy and the learned policy facilitated by A2PR.

\begin{assumption}\label{assump1}
    Supposed that $Q(s,a)$ and $P(s'|s,a)$ are Lispchitz continuous w.r.t $a$, then
        \begin{align}
            ||Q(s,a_1)-Q(s,a_2)||&\leq L_Q||a_1-a_2|| \label{eq:LQ}\\
            ||P(s'|s,a_1)-P(s'|s,a_2)||&\leq L_P||a_1-a_2|| \label{eq:LP}
        \end{align}
    for all $(s,a_1),(s,a_2)\in \mathcal{S \times A}$. $L_Q$ and $L_P$ represent the Lipschitz constants. Equation~(\ref{eq:LQ}) is frequently employed in the theoretical analysis of RL~\cite{saxena2023off,gouk2021regularisation}. Equation~(\ref{eq:LP}) has received substantial attention in theoretical RL research~\cite{dufour2013finite}.
\end{assumption}

\begin{proposition}[Behavior Policy Improvement Guarantee]
    \label{proposition2}
    Given the accurate state-action value function $Q(s,a)$, the high-advantage actions from the improved VAE and the dataset own accurate advantage $A(s,a)$. Then, we have
    \begin{equation}
    \begin{aligned}
        J(\tilde{\pi}_\beta) - J(\pi_\beta) \geq 0 ,\\
    \end{aligned}
    \end{equation}
    where denote the augmented behavior policy combined the dataset with the improved VAE as $\tilde{\pi}_\beta$, the original behavior policy of the pre-collected dataset as $\pi_\beta$.  
\end{proposition}
    The proof is provided in Appendix~\ref{appendix:A2}. Proposition~\ref{proposition2} implies that A2PR can acquire an augmented behavior policy to effectively constrain the learned policy, thereby ensuring a performance guarantee for the learned policy~\cite{hong2023beyond}.

\begin{theorem}\label{theorem1}
    With policy constraint, we have $||\pi_{\phi}(s)-\Bar{a}|| \leq \epsilon_0$. Then based on Equation (\ref{final_a}), let $||\Bar{a} - \pi_{\beta}(s)|| \leq \epsilon_1$ due to high-advantage actions from the augmented behavior policy of A2PR. With Assumption~\ref{assump1}, then we have
    \begin{equation}
    \vspace{-1mm}
    \begin{aligned}\label{eq:theorem1}
        ||Q(s,\pi_{\phi}(s)) - Q(s,\pi_{\beta}(s))|| \leq L_Q(\epsilon_0 + \epsilon_1),\\ 
    \end{aligned}
    \vspace{-1mm}
    \end{equation}
    for any $s\in \mathcal{S}.$
    \end{theorem}
    The proof is provided in Appendix~\ref{appendix:A3}.  For an accurate estimation of the true Q-function $Q^{\pi}(s',\pi(s'))$, an approximately correct estimate $\hat{Q}^\pi(s',\pi(s'))$ is required. With a sufficiently large number of samples, $\tilde{Q}^\pi(s',\pi(s'))$ will converge to $Q^{\pi}(s',\pi(s'))$, causing the overestimation error $\delta_{error}$ to approach zero~\cite{fujimoto2019off}, $\delta_{error} = \tilde{Q}^\pi(s,a) - Q^\pi(s,a)$. Both $\tilde{Q}^\pi(s',\pi(s'))$ and $Q^{\pi}(s',\pi(s'))$ satisfy Assumption~\ref{assump1}. Therefore, with Theorem~\ref{theorem1}, we have
    \begin{equation}
        ||{Q}^\pi(s',\pi_{\phi}(s')) - \tilde{Q}^\pi(s',\pi_{\beta}(s'))|| \leq 2L_Q(\epsilon_0 + \epsilon_1) + \delta_{error},
    \end{equation}
    The detailed proof is provided in Appendix~\ref{appendix:A3}. In conclusion, A2PR demonstrates its effectiveness in mitigating the problem of overestimation in value estimation.
    \begin{theorem}[Performance Gap of A2PR]\label{theorem2}
    Considering Equation (\ref{final_a}), suppose $||\bar{a} - \tilde\pi_{\beta}(s)|| \leq \tilde\epsilon_1$ and $\max_{s\in \mathcal{S}}|\pi^*(s)-\tilde\pi_{\beta}(s)| \leq \tilde\epsilon_{*}$, conditions that can be satisfied by A2PR. Then we have
    \begin{equation}
    \vspace{-1mm}
    \begin{aligned}
        |J(\pi^*) - J(\pi)| \leq \frac{\mathcal{C}L_{P}R_{max}}{1-\gamma}(\epsilon_0+\tilde\epsilon_1+\tilde\epsilon_{*}),\\ 
    \end{aligned}
    \vspace{-1mm}
    \end{equation}
    where $\mathcal{C}$ is a positive constant, and $\tilde\epsilon_1$ represents the extent of difference between the high-advantage actions and the actions from the original behavior policy.
\end{theorem}

    The detailed proof is deferred to Appendix~\ref{appendix:A4}. 
    According to Theorem~\ref{theorem2}, the performance gap is influenced by $\tilde\epsilon_1$ and $\tilde\epsilon_*$. 
    The high-advantage actions originate from both the improved VAE and the dataset.
    Considering VAE as a straightforward and effective method for learning the behavior policy~\cite{wu2022supported,zhou2021plas,fujimoto2019off}, the high-advantage actions exhibit minimal deviation from the behavior policy. 
    
    In the standard approach without the advantage-guided method, the distance between the average action $\bar a$ and the behavior policy $\pi_{\beta}(s)$ is constrained by $||\bar a-\pi_{\beta}(s)||\leq\epsilon_1$. However, in our advantage-guided approach, the modified behavior policy $\tilde\pi_{\beta}(s)$ results in a smaller distance, $||\bar a-\tilde\pi_{\beta}(s)||\leq\tilde\epsilon_1$.
    The behavior policy $\pi_{\beta}(s)$ is typically used for generating datasets like the original medium-replay dataset. In contrast, our method enhances the behavior policy by selectively incorporating data that demonstrates a higher advantage. 
    The actions produced by $\tilde\pi_{\beta}(s)$ not only exhibit a higher average advantage but also maintain a closer proximity to the high-advantage actions $\bar a$. Consequently, this leads to a reduced distance measure, where $\tilde\epsilon_1\leq\epsilon_1$. 
    This tighter bound implies that our advantage-guided method facilitates a more precise alignment with desirable actions, thereby enhancing the overall efficacy of the behavior policy.
    
    Our method optimizes this training process by focusing primarily on high-advantage data from the datasets. This approach effectively filters out less beneficial initial data, while concurrently generating a refined behavior policy, denoted as $\tilde\pi_{\beta}(s)$. This improved behavior policy is more closely aligned with the optimal policy $\pi^*(s)$, resulting in a reduced error measure, where $\tilde\epsilon_{*} \leq \epsilon_{*}$.
    In the context of the original behavior policy $\pi_{\beta}(s)$, the performance bound is given by $|J(\pi^*) - J(\pi)| \leq \frac{\mathcal{C}L_{P}R_{max}}{1-\gamma}(\epsilon_0+\epsilon_1+\epsilon_{*})$. With our advantage-guided approach, we refine this bound to $\epsilon_0 + \tilde\epsilon_1 + \tilde\epsilon_{*} \leq \epsilon_0 + \epsilon_1 + \epsilon_{*}$. This demonstrates that our method can effectively narrow the performance gap between the learned policy and the optimal policy. By selectively focusing on high-advantage data, our method not only enhances the quality of the behavior policy but also contributes to more efficient learning outcomes.

\begin{table*}[t]
    \caption{The performance of A2PR and competing baselines on D4RL datasets (Gym, AntMaze). The results for A2PR correspond to the mean and standard errors of normalized D4RL scores over the final 10 evaluations and 5 random seeds. 
    }
    \label{d4rl-result}
    \vskip 0.15in
	\centering
	\small
    \setlength{\tabcolsep}{4pt}
    \begin{tabular}{lcccccccc||c}
    \hline
	\toprule
	\multicolumn{1}{c}{\bf Task Name}  & \multicolumn{1}{c}{\bf TD3+BC} & \multicolumn{1}{c}{\bf BCQ} & \multicolumn{1}{c}{\bf BEAR} & \multicolumn{1}{c}{\bf CQL} & \multicolumn{1}{c}{\bf IQL} & \multicolumn{1}{c}{\bf AW} & \multicolumn{1}{c}{\bf OAP} &\multicolumn{1}{c}{\bf PRDC}& 
        \multicolumn{1}{c}{\bf A2PR(ours)} \\ 
	\midrule
        halfcheetah-random & $11.0$ & $8.8$ & $15.1$  & $20.0$ & $11.2$  &$16.3$& $24.0 \pm 1.6$ & $26.9\pm1.0$& $\bm{31.77\pm 0.63}$\\ 
        hopper-random & $8.5$ & $7.1$ & $14.2$ & $8.3$ & $7.9$  &$7.9$& $8.8 \pm 1.8$ & $26.8\pm9.3$& $\bm{31.55\pm 0.29}$\\ 
        walker2d-random & $1.6$ & $6.5$ & $\bm{10.7}$  & $8.3$ & $5.9$  &$4.8$& $5.1 \pm 5.1$ & $5.0\pm1.2$& $5.0\pm1.1$\\ 
        \midrule
        halfcheetah-medium & $48.3$ & $47.0$ & $41.0$  & $44.0$ & $47.4$  &$46.5$& $56.4 \pm 4.3$ &$63.5 \pm 0.9$ & $\bm{68.61 \pm 0.37}$\\ 
        hopper-medium & $59.3$ & $56.7$ & $51.9$  & $58.5$ & $66.2$  &$67.7$& $82.0 \pm 6.6$ & $100.3 \pm 0.2$ & $\bm{100.79 \pm 0.32}$  \\
        walker2d-medium & $83.7$ & $72.6$ & $80.9$  & $72.5$ & $78.3$  &$81.3$& $85.6 \pm 1.2$ & $85.2 \pm 0.4$ & $\bm{89.73 \pm 0.60}$ \\ 
        \midrule
        halfcheetah-medium-replay & $44.6$ & $40.4$ & $29.7$ & $45.5$ & $44.2$ &$44.7$ & ${53.4 \pm 1.9}$ & $55.0 \pm 1.1$ & $\bm{56.58 \pm 1.33}$\\ 
        hopper-medium-replay & $60.9$ & $53.3$ & $37.3$ & $95.0$ & $94.7$  &$97.0$& $98.5 \pm 2.5$ & ${100.1 \pm 1.6}$ & $\bm{101.54 \pm 0.90}$\\ 
        walker2d-medium-replay & $81.8$ & $52.1$ & $18.5$  & $77.2$ & $73.8$ &$78.1$ & $84.3 \pm 2.7$ &{$92.0\pm1.6$} & $\bm{94.42 \pm1.54}$\\ 
        \midrule
        halfcheetah-medium-expert & $90.7$ & $89.1$ & $38.9$  & ${91.6}$ & $86.7$  &$89.8$& $83.4 \pm 5.3$ &${94.5\pm0.5}$ & $\bm{98.25 \pm3.20}$\\ 
        hopper-medium-expert & $98.0$ & $81.8$ & $17.7$   & ${105.4}$ & $91.5$ &$104.6$ & $85.9 \pm 6.6$ &$109.2\pm4.0$ & $\bm{112.11 \pm 0.32}$ \\ 
        walker2d-medium-expert & $110.1$ & $109.5$ & $95.4$  & $108.8$ & $109.6$ &$109.4$ & ${111.1 \pm 0.6}$ &$111.2\pm0.6$ & $\bm{114.62 \pm 0.78}$\\ 
            \midrule
            {\bf{Gym Average}} & $698.5$ & $615.9$ & $450.5$ & $724.1$ & $717.5$ & $748.1$& $778.5$ &$869.5$ &$\bm{944.27}$ \\ 
        \midrule
        antmaze-umaze & $91.3$ & $0.0$ & $73.0$  & $84.8$ & $88.2$ &$77.3$& $90.4\pm5.2$ & $98.8\pm1.0$ & $\bm{99.20\pm1.60}$ \\ 
        antmaze-umaze-diverse & $54.6$ & $61.0$ & $61.0$ & $43.3$ & $66.7$ &$36.0$& $75.0\pm19.0$ & $\bm{90.0\pm6.8}$ & $84.80\pm4.49$\\ 
        antmaze-medium-play & $0.0$ & $0.0$ & $0.0$  & $65.2$ & $70.4$ &$10.7$& $62.0\pm10.0$ & $82.8\pm4.8$ & $\bm{85.60\pm9.75}$ \\ 
        antmaze-medium-diverse & $0.0$ & $0.0$ & $8.0$ & $54.0$ & $74.6$ &$6.0$& $54.5\pm23.3$ & $78.8\pm6.9$ & $\bm{85.60\pm4.63}$\\ 
        antmaze-large-play & $0.0$ & $6.7$ & $0.0$ & $18.8$ & $43.5$ &$1.3$& $0$ & $54.8\pm10.9$ & $\bm{71.20\pm5.74}$\\ 
        antmaze-large-diverse & $0.0$ & $2.2$ & $0.0$ & $31.6$ & $45.6$ &$2.0$& $9.4\pm8.4$ & $50.0\pm5.4$ & $\bm{52.80\pm9.77}$ \\ 
		\midrule
		{\bf{Antmaze Average}} & $145.9$ & $69.9$ & $142.0$ & $297.7$ & $389.0$ & $133.3$ &$291.3$ &$455.2$ &$\bm{478.2}$ \\ 
		{\bf{Total Average}} & $844.4$ & $685.8$ & $592.5$ & $1021.8$ & $1106.5$ & $881.4$ &$1069.8$ &$1324.7$ &$\bm{1422.47}$ \\ 
		\bottomrule
      \hline
	\end{tabular}
\end{table*}

\section{Experiments}\label{sec_experiment}

In this section, we begin by detailing the experimental setup in Section~\ref{setup}. Following that, we present the primary results on the D4RL benchmark dataset in Section~\ref{res_d4rl}. Subsequently, A2PR undergoes evaluation on supplementary low-quality datasets to assess its generalization capabilities in Section~\ref{eva_addition}. We then investigate its effectiveness in mitigating the overestimation issue in Section~\ref{overesti}. Lastly, we conduct a comprehensive ablation study in Section~\ref{ablation}.
\subsection{Setup}\label{setup}
    \paragraph{Datasets} We conduct our evaluations on two task domains from the D4RL benchmark \cite{fu2020d4rl}: Gym and AntMaze. All datasets used are of the "v2" version. The Gym-MuJoCo locomotion tasks serve as widely recognized standard benchmarks for assessment, encompassing three diverse environments (halfcheetah, hopper, and walker2d). These environments feature a multitude of trajectories and possess inherently smooth reward functions. The AntMaze tasks, on the other hand, present challenging scenarios with sparse rewards. These tasks require the agent to navigate through mazes, piecing together sub-optimal trajectories to reach specified goals. The AntMaze environment includes various maze layouts (umaze, medium, large), each presenting diverse challenges for an 8-DoF Ant robot.
    
    \textbf{Baselines}\quad
    We conduct a comparative analysis of our method against several robust baselines, incorporating three state-of-the-art algorithms: AW \cite{hong2023harnessing}, OAP \cite{yang2023boosting}, and PRDC \cite{ran2023policy}. AW utilizes trajectory returns to reweight the dataset for policy improvement, relying solely on the return of entire trajectories without additional interaction data. OAP introduces different policy constraints by leveraging query preferences between pre-collected and learned policy. PRDC, on the other hand, constrains the policy by searching the dataset for the nearest state-action sample. Further details on the baseline algorithms are available in Appendix~\ref{appendix:d4rl}.

\subsection{Main results on benchmark}\label{res_d4rl}

\begin{figure}[ht]
    \vspace{4mm}
    \centering
    \subfigure[]{
		\begin{minipage}[t]{0.5\linewidth}
			\centering
			\includegraphics[width=1.6in]{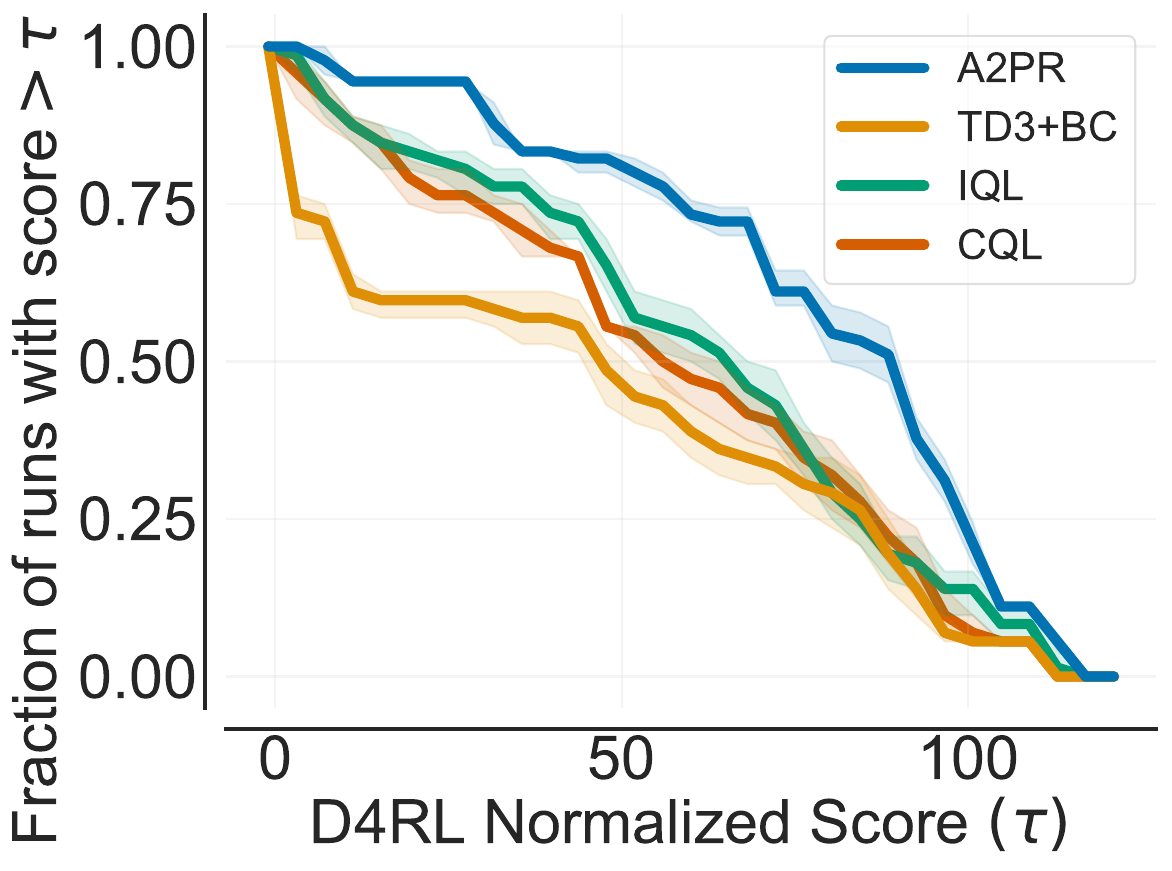}
		\end{minipage}%
	}%
	\subfigure[]{
		\begin{minipage}[t]{0.5\linewidth}
			\centering
			\includegraphics[width=1.5in]{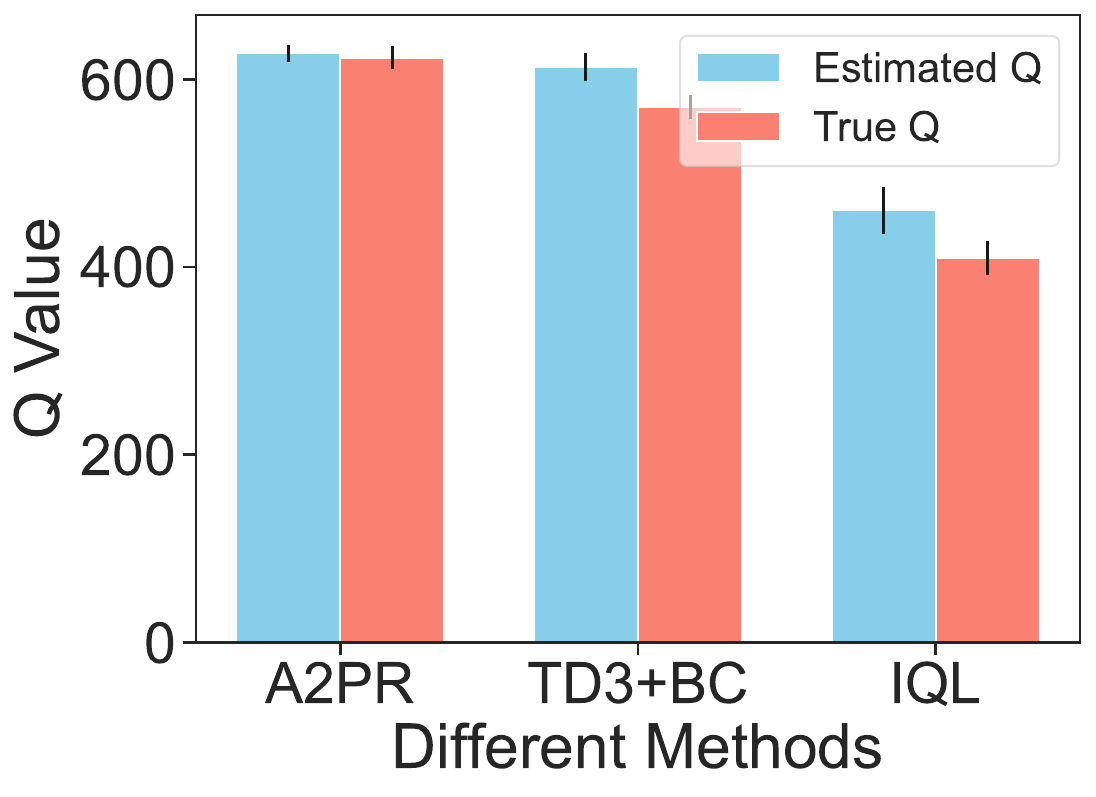}
		\end{minipage}%
	}%
    \caption{The performance profiles of reliable evaluation on D4RL based on 18 tasks and 5 random seeds for each task and the comparison between estimated Q-value and true Q-value of different methods.
    }
    \label{fig:pref_pro}
    \vspace{-5mm}
    \end{figure}

    In this section, we present the results of A2PR and competing baselines on D4RL datasets, as summarized in Table~\ref{d4rl-result}. The baseline results are directly sourced from their respective papers. For our method, A2PR is trained for 1 million steps with 5 random seeds. The experimental outcomes showcase the superiority of our approach, outperforming other baselines and achieving state-of-the-art performance in 16 out of 18 tasks. Beyond the D4RL dataset performance in Table~\ref{d4rl-result}, we provide a more comprehensive evaluation of the algorithms introduced by~\cite{tarasov2022corl}, as depicted in Figure~\ref{fig:pref_pro}(a). The statistical robustness offered by the results in Figure~\ref{fig:pref_pro}(a) complements the findings in Table ~\ref{d4rl-result}, robustly affirming the effectiveness of our method. Additional implementation details can be found in Appendix~\ref{appendix:d4rl}.

\subsection{Evaluation on additional low-quality datasets}\label{eva_addition}
\paragraph{Multiple target maze}

    We investigate a maze task with continuous actions in a 2D space. The observation comprises the agent's location and velocities, and the action is represented as $a \in [-1,1]$, indicating the linear force applied to the agent in the x and y directions. The environment features three destinations, each associated with a unique reward ($r=4,2,1$), located at $(1,1), (6,1)$, and $(1,6)$ on the map, respectively. The goal in this task is to navigate from a predefined starting location to the position that offers the highest reward.
    In this section, we formulate an offline dataset $\mathcal{D}={(s_i, a_i, r_i, d_i)}_{i=1}^{M}$ with $M=100,000$. The dataset comprises trajectories targeting different positions. Specifically, the robot trajectory dataset is designed to reach positions $(1,1)$, $(6,1)$, and $(1,6)$, accounting for 5\%, 45\%, and 50\% of the dataset, respectively.
    
\begin{figure}[ht]
    \vskip 0.1in
    \centering

    \subfigure[TD3+BC: all trajectories]{
    \includegraphics[width=3.5cm]{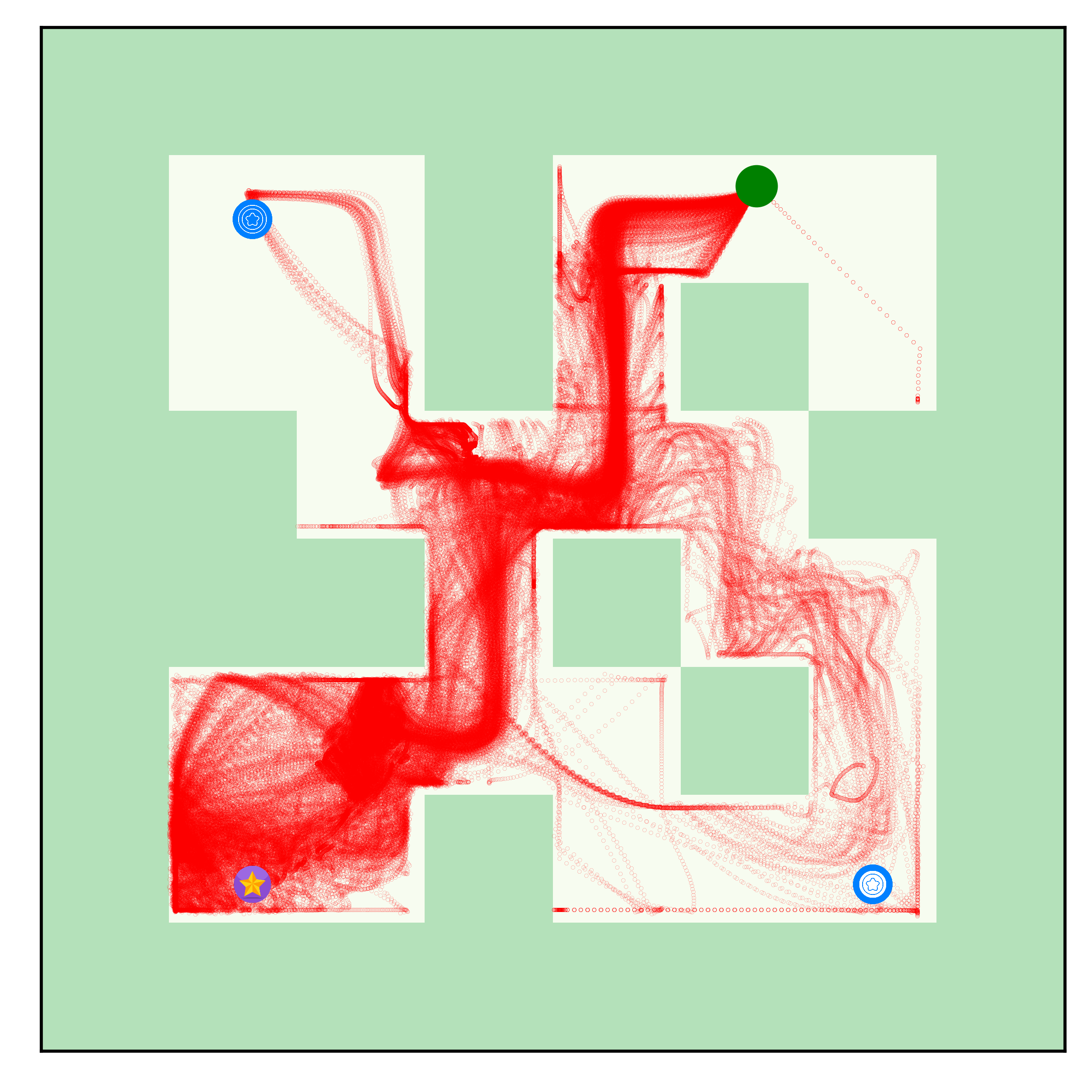}
    }
    \quad
    \subfigure[A2PR: all trajectories]{
    \includegraphics[width=3.5cm]{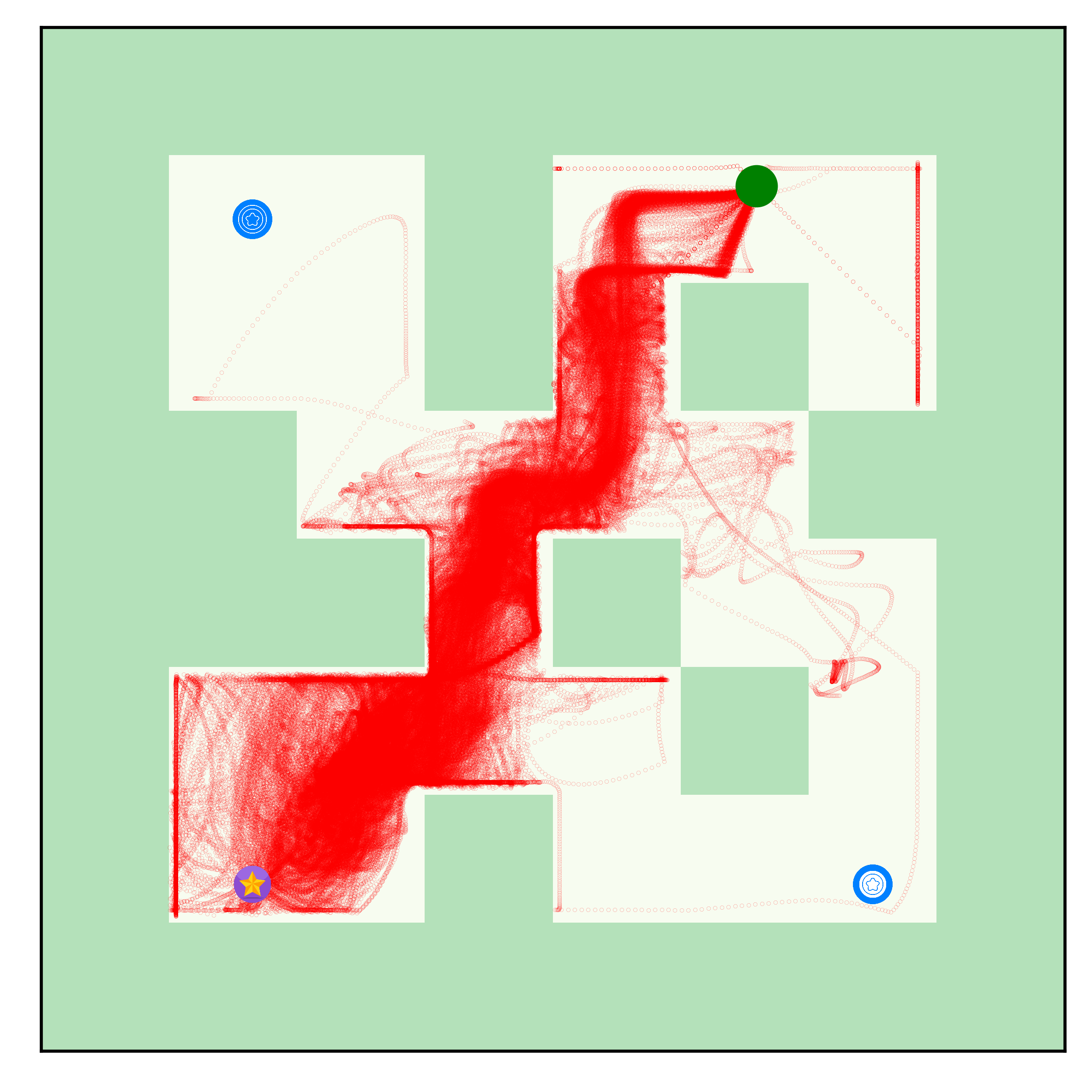}
    }

    \subfigure[TD3+BC: trajectories of final 100,000 step]{
    \includegraphics[width=3.5cm]{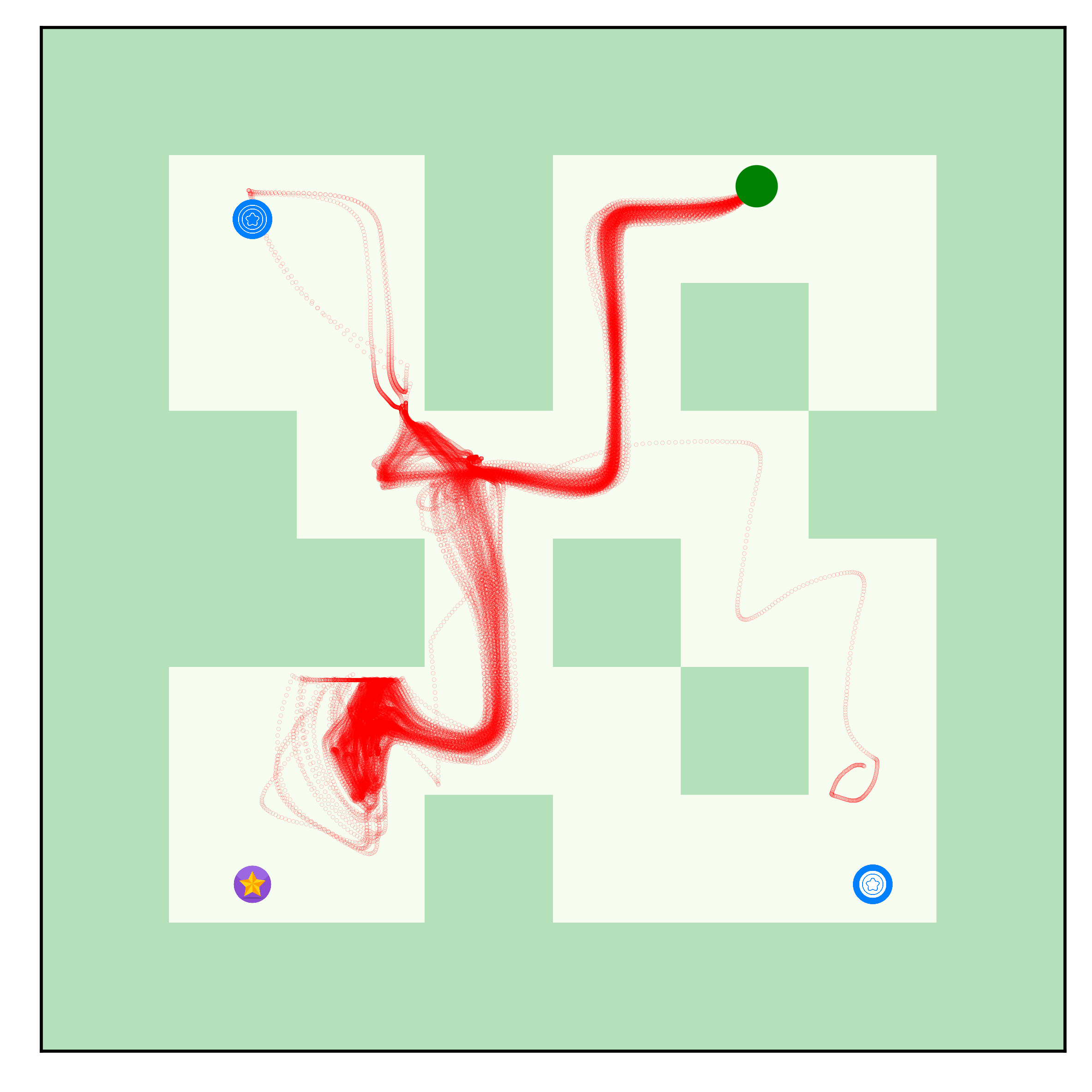}
    }
    \quad
    \subfigure[A2PR: trajectories of final 100,000 step]{
    \includegraphics[width=3.5cm]{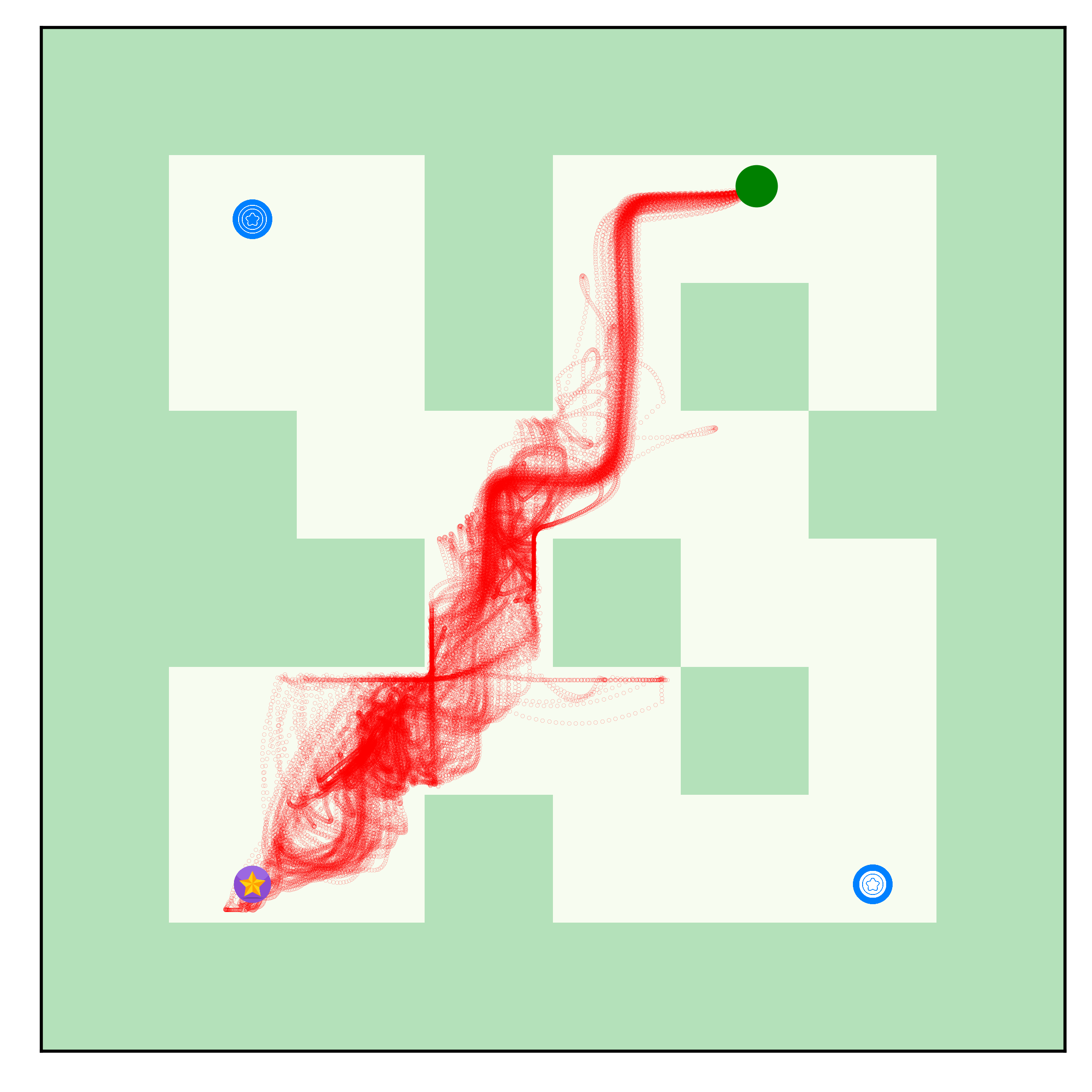}
    }
    
    \caption{All trajectories and the trajectories from the final $100,000$ steps of the trained policy for both A2PR and TD3+BC. 
    }
    \label{fig:traj}
    \vspace{-5mm}
\end{figure}

    To evaluate the effect of fixed policy constraints on low-quality trajectory datasets, we conducted a comparative study of A2PR and TD3+BC. Both methods underwent training for 500,000 steps to ensure sufficient convergence. Figure \ref{fig:traj} illustrates all trajectories as well as those from the final 100,000 steps for both A2PR and TD3+BC. Notably, the policy derived from TD3+BC demonstrates limitations, as it remains confined to suboptimal performance levels and fails to converge towards the optimal target, as highlighted in Figure \ref{fig:traj}(c). In contrast, A2PR exhibits less susceptibility to the influence of suboptimal data, successfully generating trajectories that converge on the high-return target, as depicted in Figure \ref{fig:traj}(d).
    The unnecessary conservative policy constraint in TD3+BC compels the learned policy to incorporate all actions within a given state from the fixed dataset. This constraint, combined with the behavior policy, assigns greater density to lower-quality data. Consequently, the trajectories produced by TD3+BC are relatively homogenous and repetitive. In contrast, the trajectories generated by A2PR are diverse, containing broad high-return trajectories.
    This variation primarily stems from A2PR's capacity to generate a larger number of high-advantage actions. By leveraging an enhanced Variational Autoencoder (VAE), A2PR distinguishes these actions from those in the dataset more effectively. This enhancement significantly improves the behavior policy, which in turn, provides more accurate guidance for the learned policy towards optimal actions.
    

\begin{figure}[ht]
    \vskip 0.1in
    \centering
    \subfigure[]{
		\begin{minipage}[t]{0.48\linewidth}
			\centering
			\includegraphics[width=1.5in]{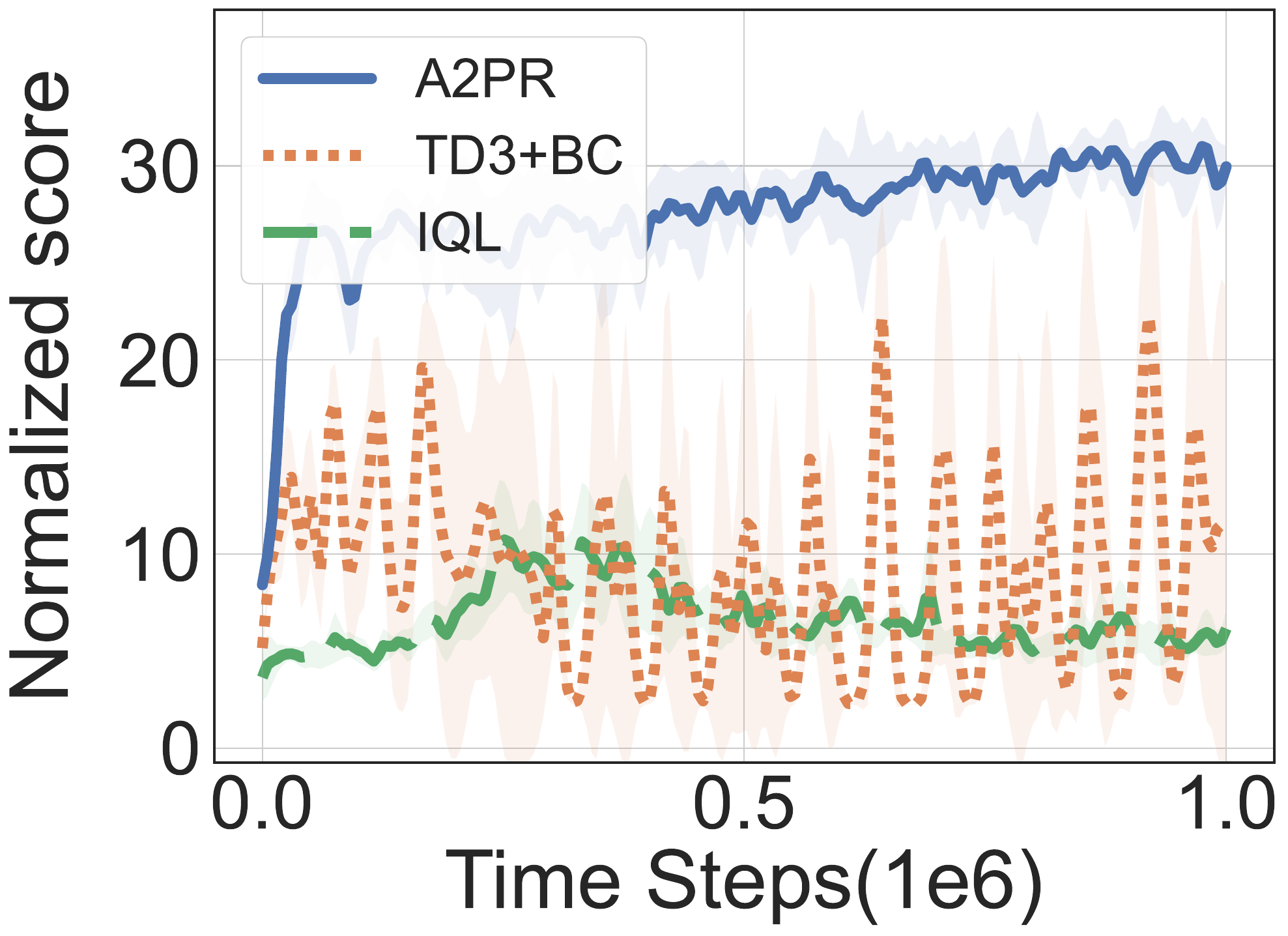}
		\end{minipage}%
	}%
	\subfigure[]{
		\begin{minipage}[t]{0.5\linewidth}
			\centering
			\includegraphics[width=1.55in]{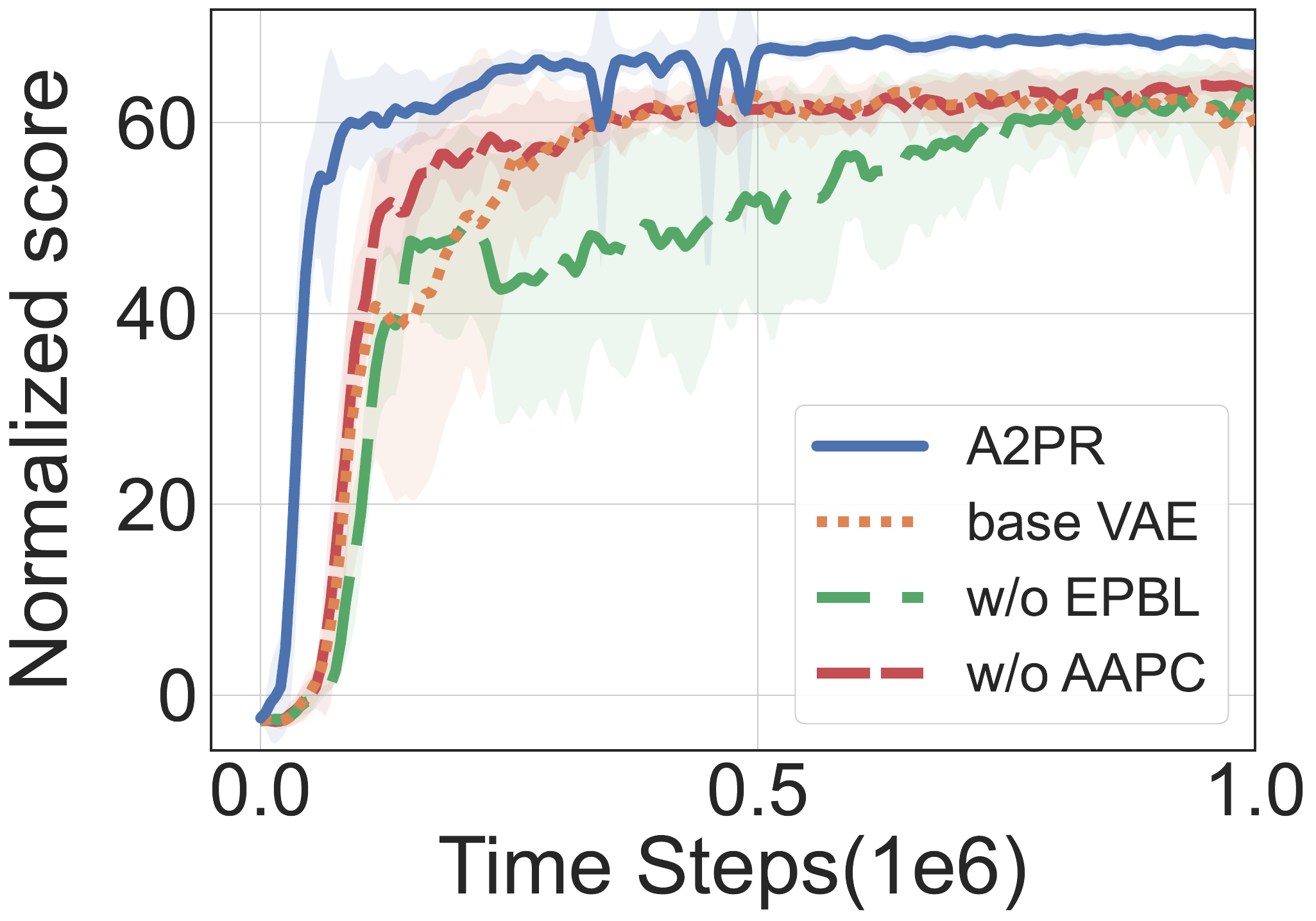}
		\end{minipage}%
	}%
    \caption{The performance of different methods in the mixed policy datasets and the comprehensive ablation study of A2PR on halfcheetah-medium-v2 with different components.
    }
    \label{fig:mix_dataset}
    \vspace{-5mm}
    \end{figure}

\paragraph{Mixed random policy lower-quality dataset}
    In this section, we aim to validate the generalization capabilities of A2PR on lower-quality datasets that consist of a substantial proportion of low-quality demonstrations. To achieve this, we evaluate A2PR alongside two strong baselines, TD3+BC and IQL, on mixed policy datasets comprising a combination of random data and expert data. The results are presented in Figure \ref{fig:mix_dataset}(a).
    The mixed policy dataset comprises 100,000 state-action pairs, mirroring the size of each task in the D4RL dataset. Comprising 99\% random policy data and 1\% expert policy data, this dataset is designed to assess the algorithms' performance on mixed-quality data. 
    
    The results indicate A2PR's superior performance compared to TD3+BC and IQL on mixed policy datasets. A substantial performance gap exists between A2PR and TD3+BC, IQL, which exhibit notably poorer performance. A2PR demonstrates improved generalization, achieving remarkable normalized scores, particularly on the halfcheetah task, even in the presence of low-quality datasets. These findings underscore A2PR's ability to mitigate the over-constraint issue associated with poorer data.

\subsection{Value estimation}\label{overesti}
    
    Value overestimation poses a significant challenge in offline RL, and we assess the comparative performance of various methods in addressing this issue using the halfcheetah-medium-v2 dataset. True Q-values are determined through Monte-Carlo rollouts~\cite{sutton2018reinforcement}. Over the 1M training steps, we randomly sample 10 states from the initial distribution, predict actions using the current policy, and interact with the environment for evaluation every 5k steps. To evaluate value estimation error, we conduct 10 final evaluations with 5 random seeds, estimating Q-values and comparing them with true Q-values across different methods. The results, depicted in Figure \ref{fig:pref_pro}(b), highlight our method's ability to achieve higher true Q-values and lower value estimation error, indicating a smaller disparity between estimated and true Q-values compared to other methods. Therefore, our proposed adaptive policy regularization approach, grounded in behavior optimization, effectively mitigates the value overestimation problem.


\subsection{Ablation Study}\label{ablation}
In this section, we perform an ablation study to assess the contributions of the main components in our algorithm. We compare normalized scores on the halfcheetah-medium-v2, as depicted in Figure~\ref{fig:mix_dataset}(b). Variants of A2PR include one without the elevated positive behavior learning (EPBL) component, denoted as w/o EPBL, and another incorporating a standard VAE instead of the improved version, referred to as base VAE. Additionally, a version of A2PR without adaptive advantage policy constraint is labeled as w/o AAPC. This analysis allows us to understand the individual impact of these components on the algorithm's performance.
 
Ablating the improved VAE leads to inferior outcomes, emphasizing the critical role of policy regularization with additional high-advantage actions from the augmented behavior policy. The convergence performance of the base VAE method experiences a slight decline, highlighting the importance of ensuring that generated actions have a higher likelihood of being high-advantage actions. Although w/o AAPC exhibits faster learning before 0.1M steps, the final performance also diminishes. This underscores the significance of selecting high-advantage actions for constraining the learned policy and guiding it toward effective policy improvement. Overall, these results underscore that A2PR achieves superior convergence performance with a swift learning pace and a high final score.

\section{Discussion}

A2PR aims to constrain the learned policy through high-advantage actions. Our initial idea aims to employ an enhanced behavior policy to restrict the learned policy, mitigating issues arising from unnecessary conservativeness towards inferior actions and preventing potential local optima or degraded policies. While using VAE appears intuitive for learning the behavior policy from the dataset, its effectiveness is hindered when the dataset itself contains more poor data. This is because VAE lacks a metric to distinguish good from bad state-action pairs during implicit variable learning, leading to an indiscriminate inclusion of all data. 
To address this, we leverage the advantage function in RL as a metric to evaluate state-action pairs' quality. Motivated by this, we combine VAE with the advantage function, using the latter to guide VAE in learning from the offline dataset with higher advantages. In a given state, the improved VAE generates higher advantage actions, and we utilize these actions, alongside those from the dataset, for policy regularization. It mitigates the issue of unnecessary conservativeness by striking a suitable balance between policy improvement and policy constraint. 

\paragraph{Limitation}
However, there remain several limitations that require further refinement in future work.
Sampling a single action using a VAE can sometimes lead to instability in learning, as the quality of a single sampled action is difficult to ensure. This variability can adversely impact the learning of the Q-function and, consequently, the policy learning process. Moving forward, selecting a more stable sampling method from the generative model will be crucial.
Therefore, we need a more accurate indicator of the quality or the stability of the sampled actions.
Additionally, the expressive capability of the behavior policy model is also important. A more expressive behavior policy model can more accurately identify high-advantage actions, thereby facilitating better policy improvement.
Therefore, having a better behavior policy to guide the learned policy is crucial in the context of offline RL. In future research endeavors, we aim to explore advanced methodologies, such as the diffusion model, a more strong expressive generative model. This exploration is geared towards achieving even more effective policy improvement.
\section{Conclusion}
\label{sec_conclusion}
    We introduce an innovative policy regularization approach, named Adaptive Advantage-Guided Policy Regularization (A2PR), designed for offline RL. To our knowledge, A2PR represents the first method to integrate VAE and the advantage function, providing a straightforward and efficient means to enhance the behavior policy. By leveraging the augmented behavior policy, A2PR effectively guides the learned policy to achieve policy improvement, mitigating the impact of suboptimal or out-of-distribution data. This approach introduces a region constraint, addressing the global constraint issues seen in prior policy regularization methods, which confined learned policies to actions within a specific state in the dataset.
    A2PR emerges as a promising solution within the realm of Offline RL, offering a robust and theoretically grounded strategy to counter unnecessary conservativeness and overestimation challenges. It attains state-of-the-art performance on the D4RL benchmark, showcasing its efficacy across diverse tasks and datasets. A2PR stands as a valuable contribution to the field, signaling potential advancements in Offline RL methodologies.


    
    

\section*{Acknowledgements}


This work is supported by the National Natural Science Foundation of China under Grant U21A20518, Grant 61825305, Grant 62102426, and Grant 62106279, for which we are immensely grateful. We would also like to thank Xianyuan Zhan, Yi-Chen Li, and the anonymous reviewers for their support and valuable discussion on this work.

\section*{Impact Statement}


This paper presents work whose goal is to advance the field of 
Machine Learning. There are many potential societal consequences 
of our work, none which we feel must be specifically highlighted here.



\bibliography{main_paper}
\bibliographystyle{icml2024}

\newpage
\appendix
\onecolumn
\section{Theoretical Proofs}
\subsection{Proof of Proposition~\ref{proposition1}}\label{appendix:A1}

    We first start with a lemma considering the behavior policy improvement as follows:
    \begin{lemma}\label{lemma1}
    Given any two policies $\pi_1$ and $\pi_2$. 
    \begin{align}
    \vspace{-1mm}
        J(\pi_1) - J(\pi_2) &= \int_s {d}_{{\pi}_1} (s) (Q_{\pi_2}(s,\pi_1(s)) - V_{\pi_2}(s))\, \mathrm{d}s \label{eq:Q}\\
        &= \int_s d_{\pi_1} (s) \int_a \pi_1(a|s) A^{\pi_2}(s, a) \, \mathrm{d}a \, \mathrm{d}s . \label{eq:A}
    \vspace{-1mm}
    \end{align}
    \end{lemma}
    \begin{proof}
    The deviation of Equation~(\ref{eq:Q}) in Lemma~\ref{lemma1} is related to~\cite{yang2023boosting,kakade2002approximately}.
    
    The deviation of Equation~(\ref{eq:A}) in Lemma~\ref{lemma1} is related to~\cite{yue2023offline}.
\end{proof}
    
    \begin{proposition}{proposition1}
    Suppose that $A^{\pi_\beta}(s,a)(\hat{\pi}_\beta(a|s)-\pi_\beta(a|s)) \geq 0$. Then, we have
    \begin{equation}
    \vspace{-1mm}
    \begin{aligned}
        J(\hat{\pi}_\beta) - J(\pi_\beta) \geq 0 ,\\ 
    \end{aligned}
    \vspace{-1mm}
    \end{equation}
    \end{proposition}
    \begin{proof}
  
    Based on Equation (\ref{elbo1}), it holds that $\forall s \in S$, $A^{\pi_\beta}(s,a)(\hat{\pi}_\beta(a|s)-\pi_\beta(a|s)) \geq 0$. 

    \begin{equation}
    \begin{aligned}
        J(\hat{\pi}_\beta) - J(\pi_\beta) &=\int_s {d}_{\hat{\pi}_\beta} (s) \int_a \hat{\pi}_\beta(a|s) A^{\pi_\beta}(s, a) \, \mathrm{d}a \, \mathrm{d}s\\
        &\geq
        \int_s {d}_{\hat{\pi}_\beta} (s) \int_a {\pi}_\beta(a|s) A^{\pi_\beta}(s, a) \, \mathrm{d}a \, \mathrm{d}s \\
        &=0. 
    \end{aligned}
\end{equation}
    {Incorporating the advantage property} $\int_a {\pi}_\beta(a|s) A^{\pi_\beta}(s, a) \, \mathrm{d}a = 0$, the above final derivation is as follows. So it follows that $J(\hat{\pi}_\beta) - J(\pi_\beta) \geq 0$. Theorem \ref{theorem1}, suggests that the preference density estimator achieves policy improvement compared to behavior policy.
\end{proof}

\subsection{Proof of Proposition~\ref{proposition2}}\label{appendix:A2}
\paragraph{Behavior Policy Improvement Guarantee.}
\begin{proof}
According to Equation~(\ref{eq:Q}) in Lemma~\ref{lemma1}, it follows that
\begin{align}
    J(\tilde{\pi}_\beta) - J(\pi) &= \int_s {d}_{\tilde{\pi}_\beta} (s) (Q(s,\tilde{\pi}_\beta(s))-V(s))\, \mathrm{d}s \label{eq:A2_1}\\
    J({\pi}_\beta) - J(\pi) &= \int_s {d}_{{\pi}_\beta} (s) (Q(s,{\pi}_\beta(s))-V(s))\, \mathrm{d}s. \label{eq:A2_2}
\end{align}
Combining Equation~(\ref{eq:A2_1}) and Equation~(\ref{eq:A2_2}), we have 
\begin{align}
    J(\tilde{\pi}_\beta) - J(\pi_\beta) &= J(\tilde{\pi}_\beta) - J(\pi) + J(\pi)- J(\pi_\beta) \\
    &= (J(\tilde{\pi}_\beta) - J(\pi)) - (J(\pi_\beta)- J(\pi)) \\
    &= \int_s {d}_{\tilde{\pi}_\beta} (s) (Q(s,\tilde{\pi}_\beta(s))-V(s))\, \mathrm{d}s - \int_s {d}_{{\pi}_\beta} (s) (Q(s,{\pi}_\beta(s))-V(s))\, \mathrm{d}s \\
    &\overset{(i)}{\approx} \int_s {d}_{{\pi}_\beta} (s) (Q(s,\tilde{\pi}_\beta(s))-Q(s,{\pi}_\beta(s)))\, \mathrm{d}s,
\end{align}
$(i)$ represents ${d}_{\tilde{\pi}_\beta} \approx {d}_{{\pi}_\beta}$ because our method only updates policies for a finite set of states in the continuous state space at each iteration, the measure of these states in the entire state space is zero. More precisely, the probability of the measure of non-overlapping states between $\tilde\pi_\beta$ and $\pi_\beta$ being zero is one. Hence, assuming that the original policy and the updated policy have approximately equal state visitation probabilities~\cite{schulman2015trust}.

Based on Equation~(\ref{eq0}), we have $A(s,\tilde{\pi}_\beta(s)) \geq A(s,{\pi}_\beta(s))$. With Equation~(\ref{eq6}), we have $Q(s,\tilde{\pi}_\beta(s)) \geq Q(s,{\pi}_\beta(s))$. Then,
\begin{equation}
\begin{aligned}
    J(\tilde{\pi}_\beta) - J(\pi_\beta) &\approx \int_s {d}_{{\pi}_\beta} (s) (Q(s,\tilde{\pi}_\beta(s))-Q(s,{\pi}_\beta(s)))\, \mathrm{d}s \\
    &\geq 0.
\end{aligned}
\end{equation}
The proof of Proposition~\ref{proposition2} is finished.
\end{proof}

\subsection{Proof of Theorem~\ref{theorem1}} \label{appendix:A3}
We first start with two lemmas as follows:
    \begin{lemma}[Triangle inequality]\label{lemma:Tri}
    For any $x\in \mathbb R^n, y\in \mathbb R^n,$, it holds that,
    \begin{equation}
        ||x+y||\leq||x||+||y||.
    \end{equation}
    \end{lemma}
    \begin{lemma}\label{lemma:3}
    With Assumption~\ref{assump1}, it holds that,
    \begin{equation}
        \int_s |{d}_{{\pi}_\phi}(s)-{d}_{{\pi}_\beta}(s)| \, \mathrm{d}s \leq \mathcal{C}L_P\max_{s\in\mathcal{S}}||\pi(s)-\pi_\beta(s)||,
    \end{equation}
    where $\mathcal{C}$ is a positive constant.

    \end{lemma}
     \begin{proof}
    The proof of Lemma~\ref{lemma:3} can be found in the appendix of \cite{xiong2022deterministic}. 
        
    Next, we will provide the proof of Theorem~\ref{theorem1}.

    Based on Lemma~\ref{lemma:Tri} and Equation (\ref{final_a}), the left side of Equation (\ref{eq:theorem1}) can be expanded as below
    \begin{equation}
    \begin{aligned}
        ||Q(s,\pi_{\phi}(s)) - Q(s,\pi_{\beta}(s))|| &= ||Q(s,\pi_{\phi}(s)) - Q(s,\Bar{a}) + Q(s,\Bar{a}) + Q(s,\pi_{\beta}(s))|| \\
        &\leq ||Q(s,\pi_{\phi}(s)) - Q(s,\Bar{a}) || + ||Q(s,\Bar{a}) + Q(s,\pi_{\beta}(s))|| \\
        &\leq L_Q(||\pi_{\phi}(s)-\Bar{a}|| + ||\Bar{a}-\pi_{\beta}(s)||) \\
        &\leq L_Q(\epsilon_0 + \epsilon_1).
    \end{aligned}
    \end{equation}
    The proof of Theorem~\ref{theorem1} is finished.
\end{proof}

\begin{proof} Next, we will demonstrate that A2PR effectively mitigates the value overestimation issue. 

With the overestimation error~\cite{fujimoto2019off} and Assumption~\ref{assump1}, then we have
    \begin{align}
        \delta_{error} = \tilde{Q}^\pi(s',\pi_{\beta}(s')) - Q^\pi(s',\pi_{\beta}(s')), \label{overesti1}\\
        ||Q^\pi(s',\pi_{\phi}(s')) - Q^\pi(s',\pi_{\beta}(s'))|| \leq L_Q(\epsilon_0 + \epsilon_1), \label{overesti2}\\
        ||\tilde{Q}^\pi(s',\pi_{\phi}(s')) - \tilde{Q}^\pi(s',\pi_{\beta}(s'))|| \leq L_Q(\epsilon_0 + \epsilon_1). \label{overesti3}
    \end{align}
    Combining Equation~(\ref{overesti1}), Equation~(\ref{overesti2}) and Equation~(\ref{overesti3}), then with Lemma~\ref{lemma:Tri} we have that
    \begin{equation}
        \begin{aligned}
            ||\tilde{Q}^\pi(s',\pi_{\phi}(s')) - Q^\pi(s',\pi_{\phi}(s'))|| &= ||\tilde{Q}^\pi(s',\pi_{\phi}(s')) - \tilde{Q}^\pi(s',\pi_{\beta}(s')) + \tilde{Q}^\pi(s',\pi_{\beta}(s'))- Q^\pi(s',\pi_{\phi}(s'))||\\
            &= ||\tilde{Q}^\pi(s',\pi_{\phi}(s')) - \tilde{Q}^\pi(s',\pi_{\beta}(s')) + Q^\pi(s',\pi_{\beta}(s')) + \delta_{error} - Q^\pi(s',\pi_{\phi}(s'))||\\
            &\leq ||\tilde{Q}^\pi(s',\pi_{\phi}(s')) - \tilde{Q}^\pi(s',\pi_{\beta}(s'))|| + ||Q^\pi(s',\pi_{\phi}(s')) - Q^\pi(s',\pi_{\beta}(s'))|| + \delta_{error}\\
            &\leq 2L_Q(\epsilon_0 + \epsilon_1) + \delta_{error}.
        \end{aligned}
    \end{equation}
    The proof of mitigating the value overestimation issue has been completed.
\end{proof}

\subsection{Proof of Theorem~\ref{theorem2}} \label{appendix:A4}

\begin{proof}
    
    With Lemma~\ref{lemma:Tri}, it follows that
    \begin{equation}\label{eq:pi_d0}
    \begin{aligned}
        |J(\pi^*) - J(\pi)| &= |J(\pi^*) -J(\tilde\pi_{\beta}) + J(\tilde\pi_{\beta}) - J(\pi)| \\
        &\leq |J(\pi^*) -J(\tilde\pi_{\beta})| + |J(\pi) - J(\tilde\pi_{\beta})|.
    \end{aligned}
    \end{equation}
    Firstly, considering $|J(\pi) - J(\tilde\pi_{\beta})|$ and Lemma~\ref{lemma:3}, we have
    \begin{equation}\label{eq:pi_d1}
    \begin{aligned}
        |J(\pi) - J(\tilde\pi_{\beta})| &= |\frac{1}{1-\gamma}\mathbb E_{s\sim d_{\pi_{\phi}}}[r(s)]-\frac{1}{1-{\gamma}}\mathbb E_{s\sim d_{\tilde\pi_{\beta}}}[r(s)]| \\
        &= \frac{1}{1-\gamma}|\int_s ({d}_{{\pi}_\phi}(s)-{d}_{\tilde\pi_{\beta}}(s)) r(s)\, \mathrm{d}s| \\
        &\leq \frac{1}{1-\gamma}\int_s |{d}_{{\pi}_\phi}(s)-{d}_{\tilde\pi_{\beta}}(s)| |r(s)| \, \mathrm{d}s \\
        &\leq \frac{R_{max}}{1-\gamma}\int_s |{d}_{{\pi}_\phi}(s)-{d}_{\tilde\pi_{\beta}}(s)|\, \mathrm{d}s \\
        &\leq \frac{\mathcal{C} L_P R_{max}}{1-\gamma} \max_{s\in S}||\pi(s)-\tilde\pi_{\beta}(s)|| \\
        &= \frac{\mathcal{C} L_P R_{max}}{1-\gamma} \max_{s\in S}||\pi(s)-\Bar{a} + \Bar{a}-\tilde\pi_{\beta}(s)|| \\
        &\leq \frac{\mathcal{C} L_P R_{max}}{1-\gamma}(||\pi(s)-\Bar{a}|| + ||\Bar{a}-\tilde\pi_{\beta}(s)||) \\
        &\leq \frac{\mathcal{C} L_P R_{max}}{1-\gamma} (\epsilon_0+\tilde\epsilon_1). \\
    \end{aligned}
    \end{equation}

Then, considering $| J(\pi^*) - J(\tilde\pi_{\beta})|$ and Lemma~\ref{lemma:3}, we get
    \begin{equation}\label{eq:pi_d2}
    \begin{aligned}
        | J(\pi^*) - J(\tilde\pi_{\beta})| &= |\frac{1}{1-\gamma}\mathbb E_{s\sim d_{\pi_{\phi}^*}}[r(s)]-\frac{1}{1-{\gamma}}\mathbb E_{s\sim d_{\tilde\pi_{\beta}}}[r(s)] | \\
        &= \frac{1}{1-\gamma}|\int_s ({d}_{{\pi}_\phi^*}(s)-{d}_{\tilde\pi_{\beta}}(s)) r(s)\, \mathrm{d}s| \\
        &\leq \frac{1}{1-\gamma}\int_s |{d}_{{\pi}_\phi^*}(s)-{d}_{\tilde\pi_{\beta}}(s)| |r(s)| \, \mathrm{d}s \\
        &\leq \frac{R_{max}}{1-\gamma}\int_s |{d}_{{\pi}_\phi^*}(s)-{d}_{\tilde\pi_{\beta}}(s)|\, \mathrm{d}s \\
        &\leq \frac{\mathcal{C} L_P R_{max}}{1-\gamma} \max_{s\in S}||\pi^*(s)-\tilde\pi_{\beta}(s)|| \\
        &\leq \frac{\mathcal{C} L_P R_{max}}{1-\gamma} \tilde\epsilon_{*}.
    \end{aligned}
    \end{equation}
Finally, combining Equation~(\ref{eq:pi_d1}) and Equation~(\ref{eq:pi_d2}), we have that
    \begin{equation}
    \begin{aligned}
        |J(\pi^*) - J(\pi)| &= |J(\pi^*) -J(\tilde\pi_{\beta}) + J(\tilde\pi_{\beta}) - J(\pi)| \\
        &\leq |J(\pi^*) -J(\tilde\pi_{\beta})| + |J(\tilde\pi_{\beta}) - J(\pi)|\\
        &\leq \frac{\mathcal{C} L_P R_{max}}{1-\gamma}(\epsilon_0+\tilde\epsilon_1+\tilde\epsilon_{*}).
    \end{aligned}
    \end{equation}
    The proof is finished. When not using the advantage-guided method, $||\bar a-\pi_{\beta}(s)||\leq\epsilon_1$.
     $\tilde\pi_{\beta}(s )$ produces actions that have a higher average advantage and a relatively smaller difference with actions $\bar a$ than the behavior policy $\pi_{\beta}(s)$, then $\tilde\epsilon_1\leq\epsilon_1$. 
     Our method selects data with higher advantage through advantage-guided, which is equivalent to using a better behavior policy $\tilde\pi_{\beta}(s)$ for generating the data.
     Thus this better behavior policy $\tilde\pi_{\beta}(s)$ reduces the error with respect to the optimal policy $\pi^*(s)$ than the behavior policy $\pi_{\beta}(s)$, so $\tilde\epsilon_{*} \leq\epsilon_{*}$. For $\pi_{\beta}(s)$, $|J(\pi^*) - J(\pi)| \leq \frac{\mathcal{C}L_{P}R_{max}}{1-\gamma}(\epsilon_0+\epsilon_1+\epsilon_{*})$. Then, $\epsilon_0+\tilde\epsilon_1+\tilde\epsilon_{*} \leq \epsilon_0+\epsilon_1+\epsilon_{*}$.Thus, our advantage-guided method can reduce this performance gap. 

\end{proof}
\begin{figure*}[ht]
\vskip 0.2in
\centering
\subfigure{
		\begin{minipage}[t]{0.33\linewidth}
			\centering
			\includegraphics[width=2.2in]{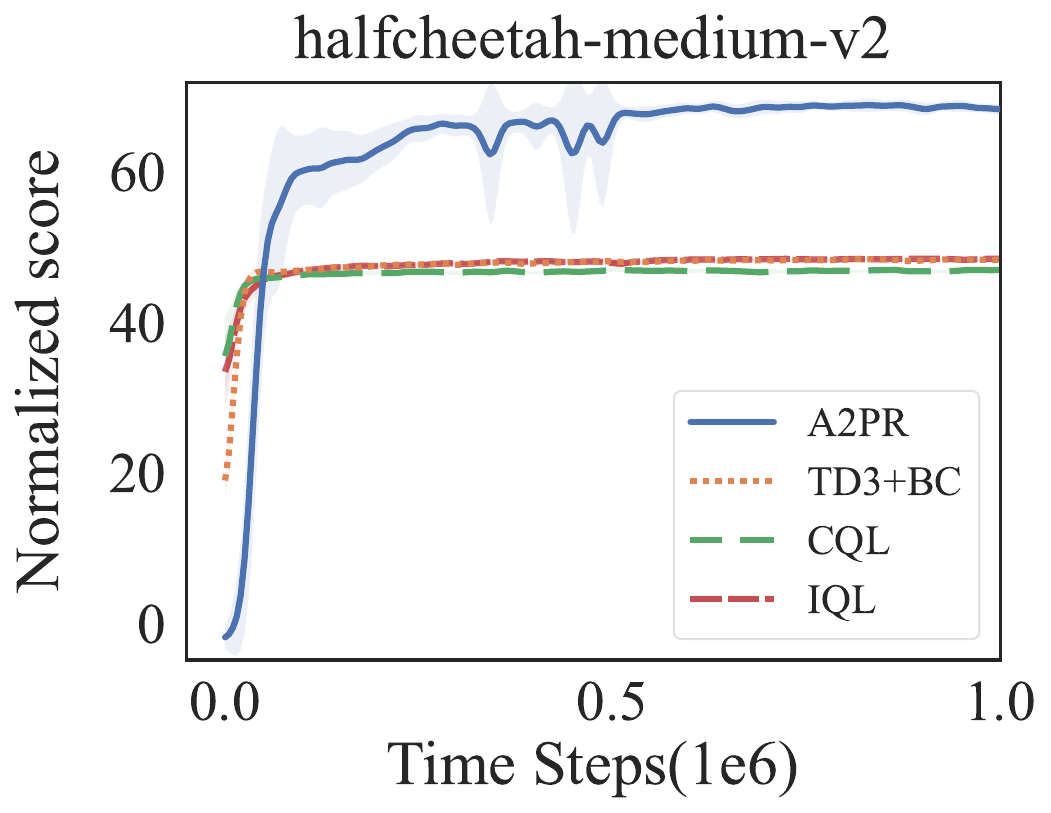}
		\end{minipage}%
	}%
\subfigure{
		\begin{minipage}[t]{0.33\linewidth}
			\centering
			\includegraphics[width=2.2in]{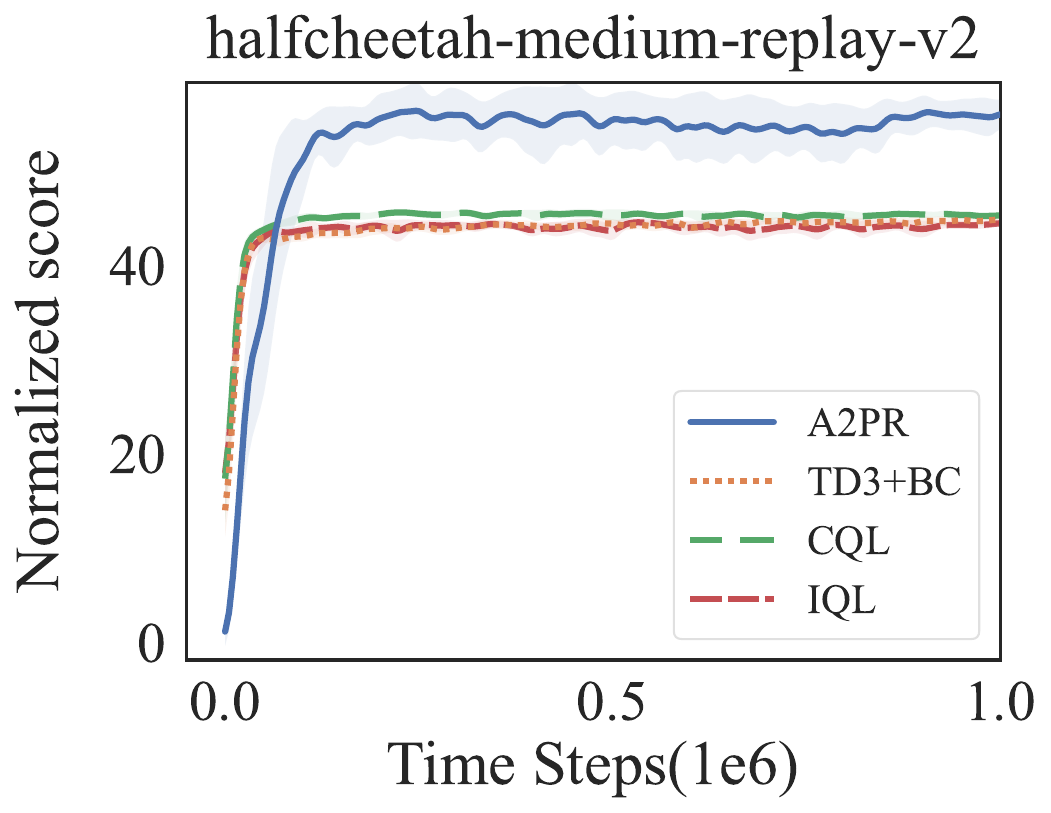}
		\end{minipage}%
	}%
\subfigure{
		\begin{minipage}[t]{0.33\linewidth}
			\centering
			\includegraphics[width=2.2in]{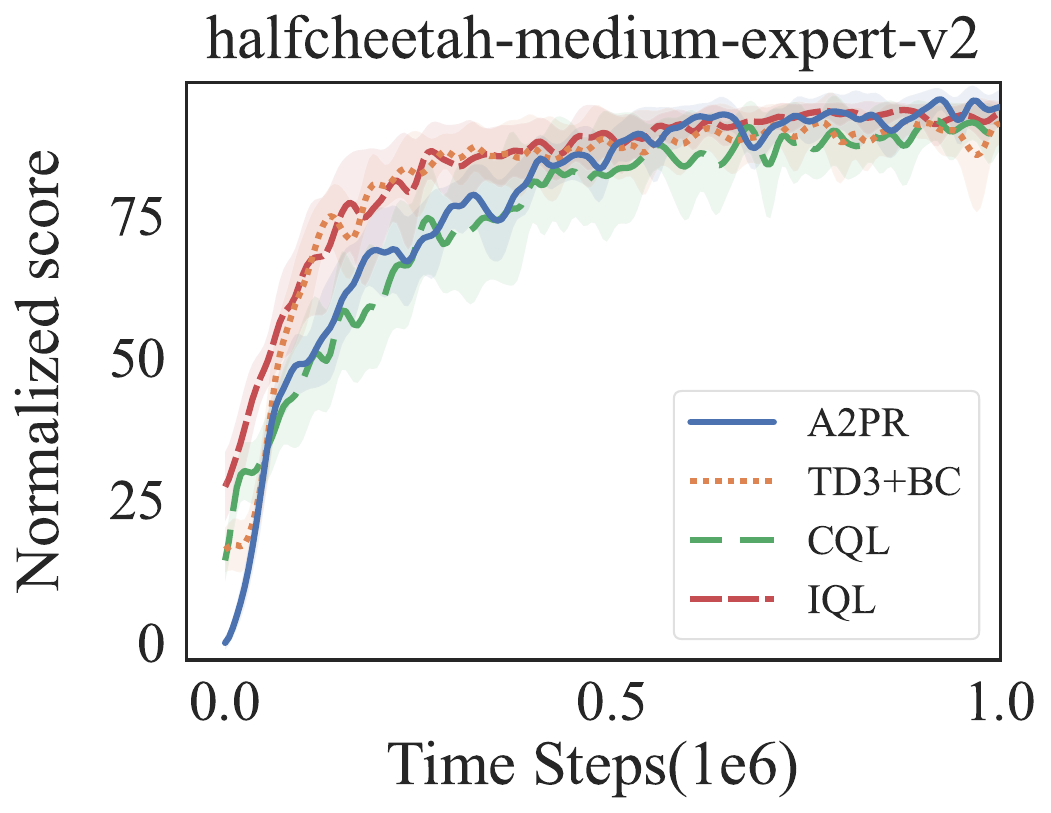}
		\end{minipage}%
	}%
\vspace{-.1in}

\subfigure{
		\begin{minipage}[t]{0.33\linewidth}
			\centering
			\includegraphics[width=2.2in]{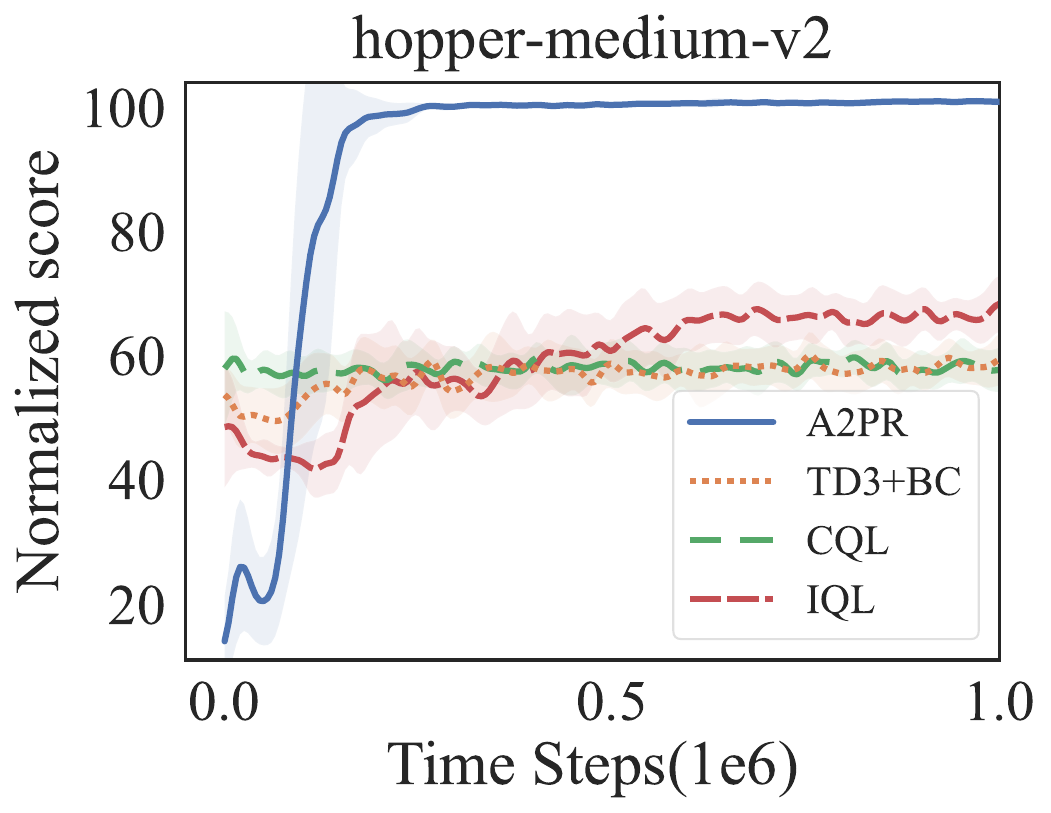}
		\end{minipage}%
	}%
\subfigure{
		\begin{minipage}[t]{0.33\linewidth}
			\centering
			\includegraphics[width=2.2in]{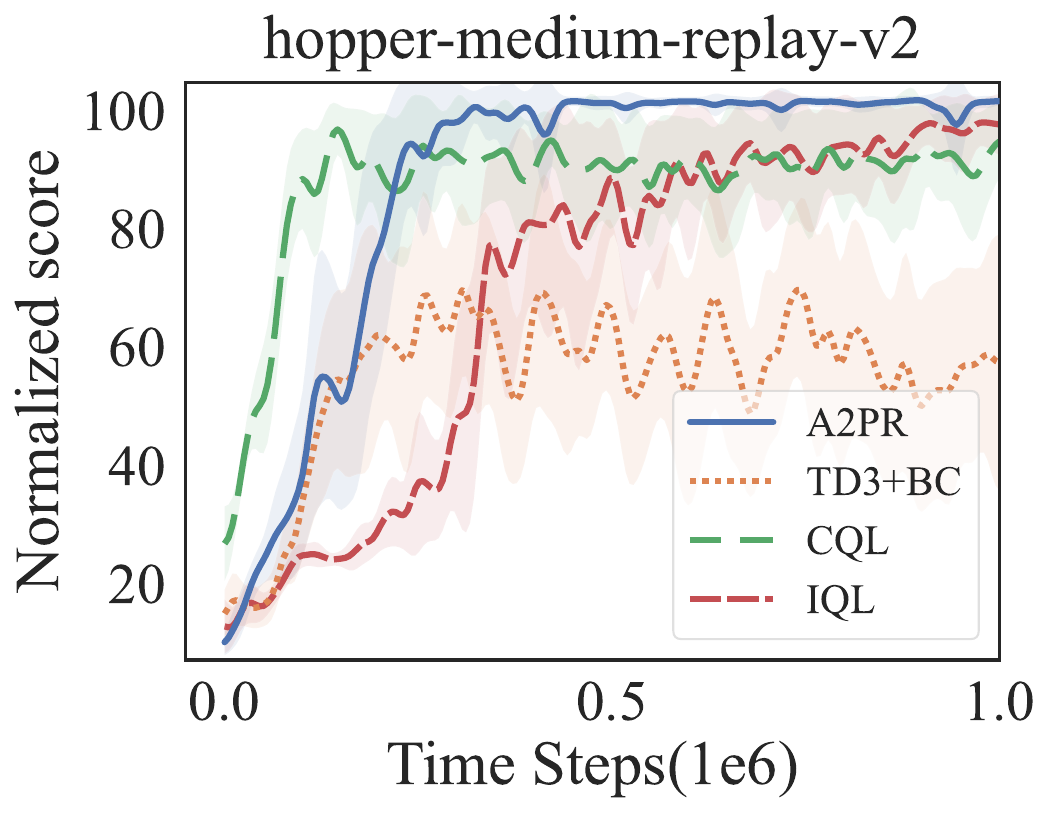}
		\end{minipage}%
	}%
\subfigure{
		\begin{minipage}[t]{0.33\linewidth}
			\centering
			\includegraphics[width=2.2in]{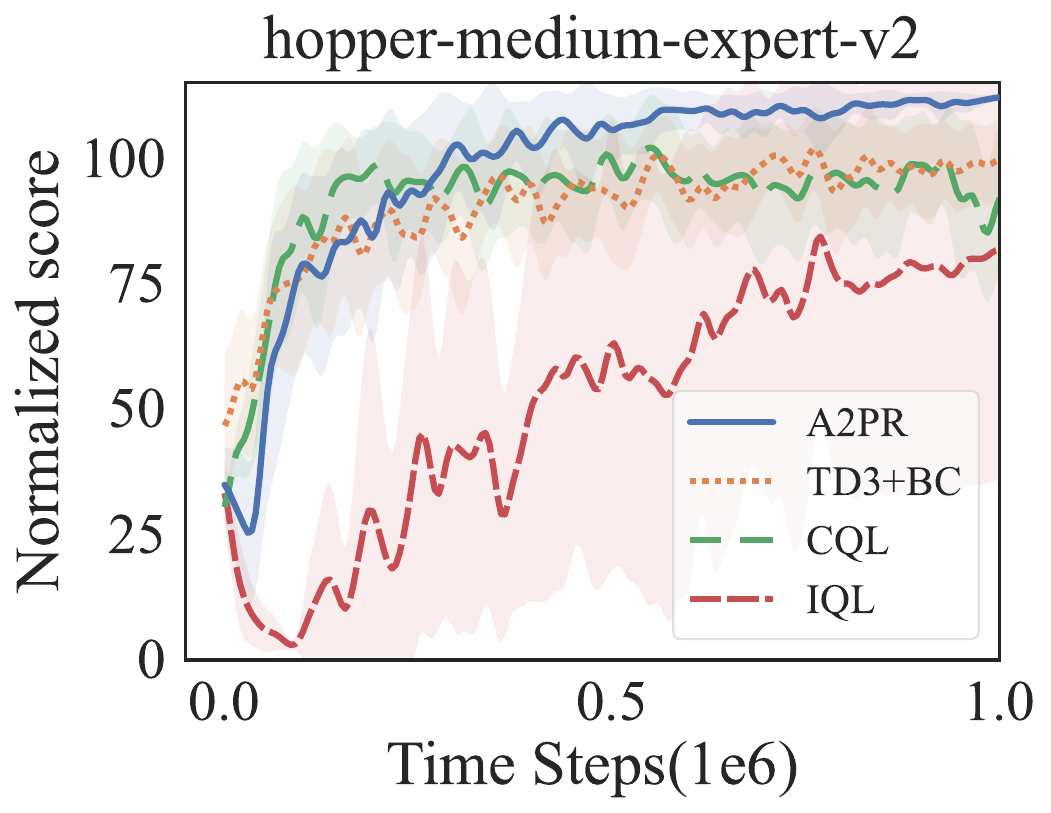}
		\end{minipage}%
	}%
 \vspace{-.1in}

\subfigure{
		\begin{minipage}[t]{0.33\linewidth}
			\centering
			\includegraphics[width=2.2in]{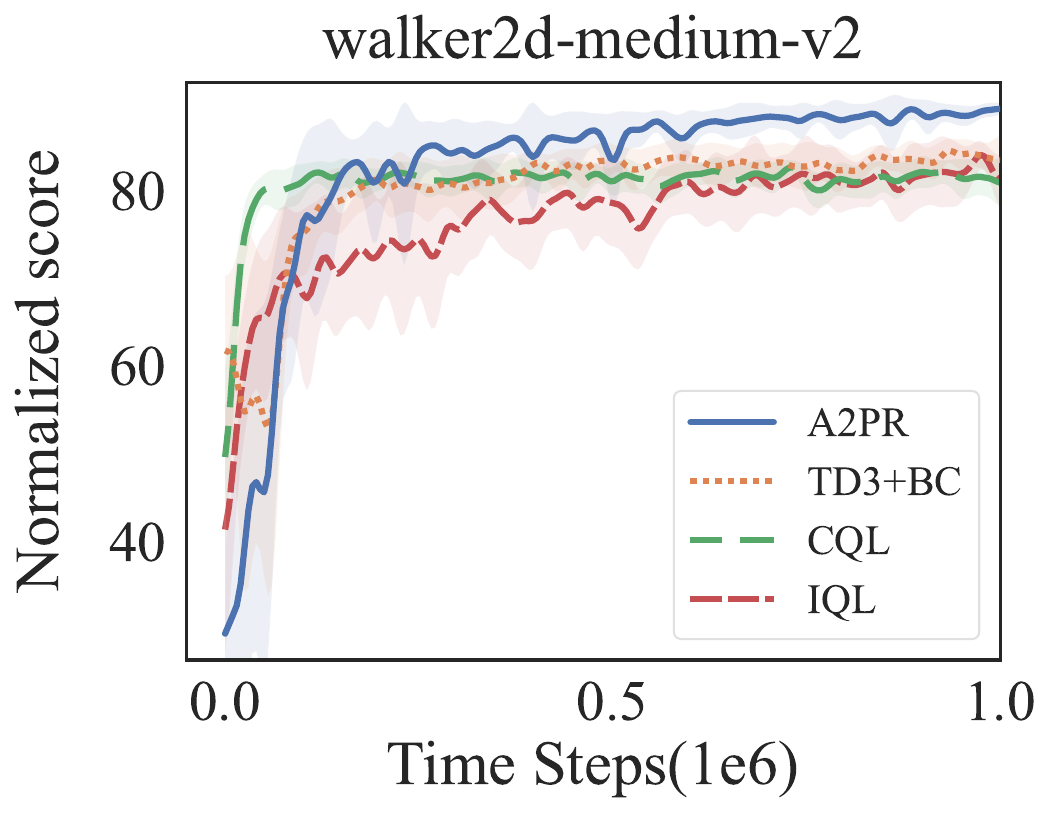}
		\end{minipage}%
}%
\subfigure{
		\begin{minipage}[t]{0.33\linewidth}
			\centering
			\includegraphics[width=2.2in]{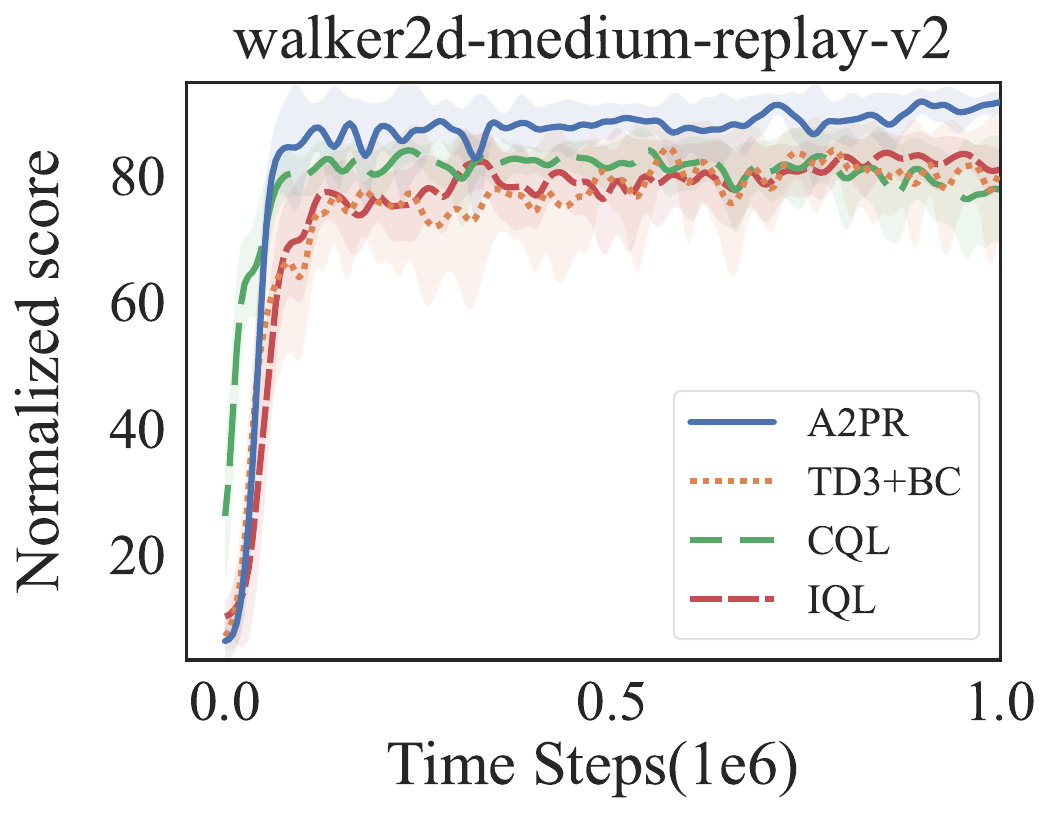}
		\end{minipage}%
	}%
\subfigure{
		\begin{minipage}[t]{0.33\linewidth}
			\centering
			\includegraphics[width=2.2in]{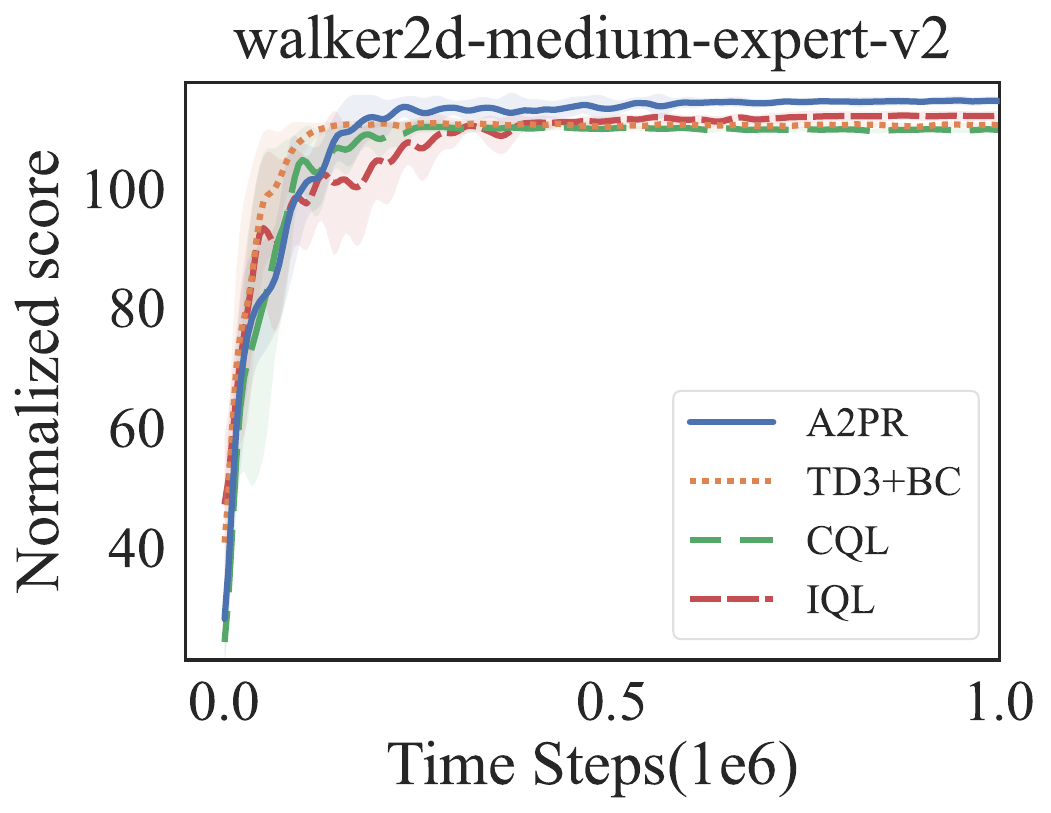}
		\end{minipage}%
	}%
\centering
\vspace{-.1in}
\caption{Results of performance comparisons conducted on nine original tasks in the D4RL dataset. The lines and shaded areas indicate the averages and standard deviations calculated over 5 random seeds, respectively.}
\vskip -0.1in
\label{fig_dmc_results}
\end{figure*}
\section{More Results}
\subsection{Main results on benchmark}\label{appendix:d4rl}
\paragraph{Baselines} We compare our method with several strong baselines, including three state-of-the-art algorithms: AW \cite{hong2023harnessing}, OAP \cite{yang2023boosting}, and PRDC \cite{ran2023policy}. Additionally, we consider policy regularization methods using behavior cloning, such as TD3+BC~\cite{fujimoto2021minimalist}; methods employing other divergences like BCQ~\cite{fujimoto2019off} and BEAR~\cite{kumar2019stabilizing} based on maximum mean discrepancy (MMD) and Gaussian kernel; Q-value constraint or critic penalty methods like CQL~\cite{kumar2020conservative}, which lower-bounds the policy's true value with a conservative Q-value function; and implicit Q learning with expectile regression, avoiding queries to values of OOD actions as in IQL~\cite{kostrikov2021offline}.
\begin{figure}[ht]
\vskip 0.1in
    \begin{center}
    \centerline{\includegraphics[width=\columnwidth]{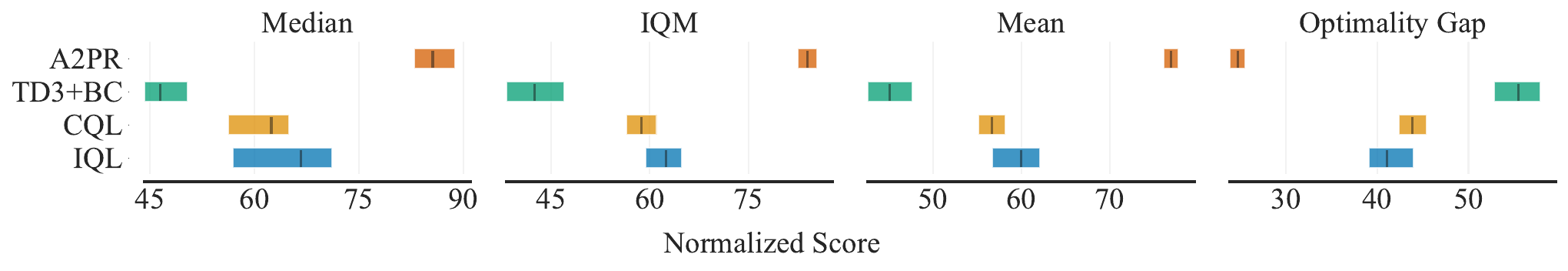}}
    \caption{Reliable evaluation for statistical uncertainty on D4RL with 95\% CIs based on 18 tasks and 5 random seeds for each task.
    }
    \label{fig:IQM}
    \end{center}
    \vskip -0.3in
\end{figure}

    In addition to presenting the D4RL dataset performance in Table~\ref{d4rl-result}, we provide a more thorough evaluation of the algorithms implemented by~\cite{tarasov2022corl}, depicted in Figure~\ref{fig:pref_pro}(a). The training curves of A2PR are compared with TD3+BC, CQL, and IQL, and the results are illustrated in Figure~\ref{fig_dmc_results}.
    Leveraging metrics from a reputable source~\cite{agarwal2021deep} enhances result confidence by addressing statistical uncertainty across multiple runs. Improved outcomes are indicated by higher mean, median, and IQM scores, along with a lower optimality gap, as illustrated in Figure~\ref{fig:IQM}. The results robustly confirm the superiority of our method.

\subsection{Sensitivity on the different advantage thresholds $\epsilon_A$}
This section explores the impact of varying advantage thresholds, denoted as $\epsilon_A$, on the performance of our policy. We conduct experiments using the A2PR algorithm across three datasets: Hopper, HalfCheetah, and Walker2d (specifically -medium-replay-v2, -medium-v2, -medium-expert-v2). For each dataset, the model was trained for 1 million steps across five different seeds, as shown in Table \ref{table:thresholds}. We assessed the performance of the policy with advantage thresholds set at $\epsilon_A \in \{-0.5, -0.1, 0, 0.1, 0.5\}$.
The results indicate that the optimal performance is achieved when $\epsilon_A = 0$. This setting allows for a balanced approach to action selection, effectively avoiding the pitfalls of high thresholds that may exclude potentially beneficial actions due to their sparse occurrence, as well as low thresholds that might include more suboptimal actions. Essentially, a zero threshold maintains a healthy balance, enabling the selection of actions that contribute positively to learning outcomes without compromising the robustness of the algorithm. Furthermore, the consistent performance across varying thresholds suggests that our algorithm is robust to changes in $\epsilon_A$. This adaptability underscores the utility of A2PR in diverse settings, making it a reliable choice for applications requiring a stable learning process.

\begin{table*}[t]
    \caption{The influence of different advantage thresholds $\epsilon_A$ on performance of D4RL datasets. The results for A2PR correspond to the mean and standard errors of normalized D4RL scores over 5 random seeds. 
    }
    \label{table:thresholds}
    \vskip 0.15in
	\centering
	\small
    \setlength{\tabcolsep}{8pt}
    \begin{tabular}{lcccccc}
    \hline
	\toprule
	\multicolumn{1}{c}{\bf Task Name}  & \multicolumn{1}{c}{$\epsilon_A$ = - 0.5} & \multicolumn{1}{c}{$\epsilon_A$ = - 0.1} & \multicolumn{1}{c}{$\epsilon_A$ = 0} & \multicolumn{1}{c}{$\epsilon_A$ = 0.1} & \multicolumn{1}{c}{$\epsilon_A$ = 0.5} &  \\ 
	\midrule
        Halfcheetah-medium-v2 & 67.45 $\pm$ 0.674 & 67.03 $\pm$ 2.56 & 68.61 $\pm$ 0.37 & \textbf{69.66 $\pm$ 0.52} & 65.71 $\pm$ 6.27 \\[2pt]
    Hopper-medium-v2 & 100.1 $\pm$ 0.88 & 99.88 $\pm$ 0.07 & \textbf{100.79 $\pm$ 0.32} & 100.1 $\pm$ 2.21 & 95.76 $\pm$ 8.27 \\[2pt]
    Walker2d-medium-v2 & 76.79 $\pm$ 7.72 & 82.41 $\pm$ 3.77 & 89.73 $\pm$ 0.60 & 88.9 $\pm$ 0.62 & \textbf{90.07 $\pm$ 3.30} \\[2pt]
     \midrule
    Halfcheetah-medium-replay-v2 & 42.80 $\pm$ 0.89 & 52.12 $\pm$ 1.71 & \textbf{56.58 $\pm$ 1.33} & 52.58 $\pm$ 3.46 & 52.72 $\pm$ 0.75 \\[2pt]
    Hopper-medium-replay-v2 & \textbf{101.7 $\pm$ 0.75} & 100.9 $\pm$ 0.5 & 101.54 $\pm$ 0.90 & 99.55 $\pm$ 1.86 & 101.2 $\pm$ 0.37 \\[2pt]
    Walker2d-medium-replay-v2 & 95.92 $\pm$ 1.13 & 90.99 $\pm$ 7.56 & 94.42 $\pm$ 1.54 & 88.22 $\pm$ 2.83 & \textbf{96.31 $\pm$ 1.96} \\[2pt]
   \midrule

    Halfcheetah-medium-expert-v2 & 87.06 $\pm$ 5.91 & 97.57 $\pm$ 2.30 & \textbf{98.25 $\pm$ 3.20} & 93.29 $\pm$ 4.38 & 93.5 $\pm$ 6.06 \\[2pt]
    Hopper-medium-expert-v2 & 112.1 $\pm$ 0.28 & 107.54 $\pm$ 2.26 & \textbf{112.11 $\pm$ 0.32} & 105.35 $\pm$ 4.38 & 96.44 $\pm$ 4.58 \\[2pt]
    Walker2d-medium-expert-v2 & 110.2 $\pm$ 2.55 & 112.42 $\pm$ 0.87 & \textbf{114.62 $\pm$ 0.78} & 105 $\pm$ 6.27 & 112.4 $\pm$ 1.25 \\ [2pt]
		\bottomrule
      \hline
	\end{tabular}
\end{table*}

\subsection{Mean advantage based on the same state}
    A2PR aims to utilize high-advantage actions to adaptively constrain the learned policy. To examine whether A2PR has learned actions with higher advantages from the low-quality dataset, A2PR, TD3+BC, and IQL are evaluated on the same 1,000 states randomly sampled from the halfcheetah-medium task dataset with 5 random seeds. Comparisons among the mean advantage curves of the actions from different methods are shown in Figure \ref{fig:compar_adv}(a). The results demonstrate that our method selects actions with higher advantages based on the same states compared to TD3+BC and IQL. Moreover, the mean advantage from our method is positive in all 1000 states with 5 random seeds. These findings provide additional confirmation that A2PR has successfully acquired advantageous actions, even when exposed to a low-quality dataset.

\subsection{A2PR implementation based on SAC framework}
\begin{figure}[ht]
    \centering
    \subfigure[]{
		\begin{minipage}[t]{0.5\linewidth}
			\centering
			\includegraphics[width=3in]{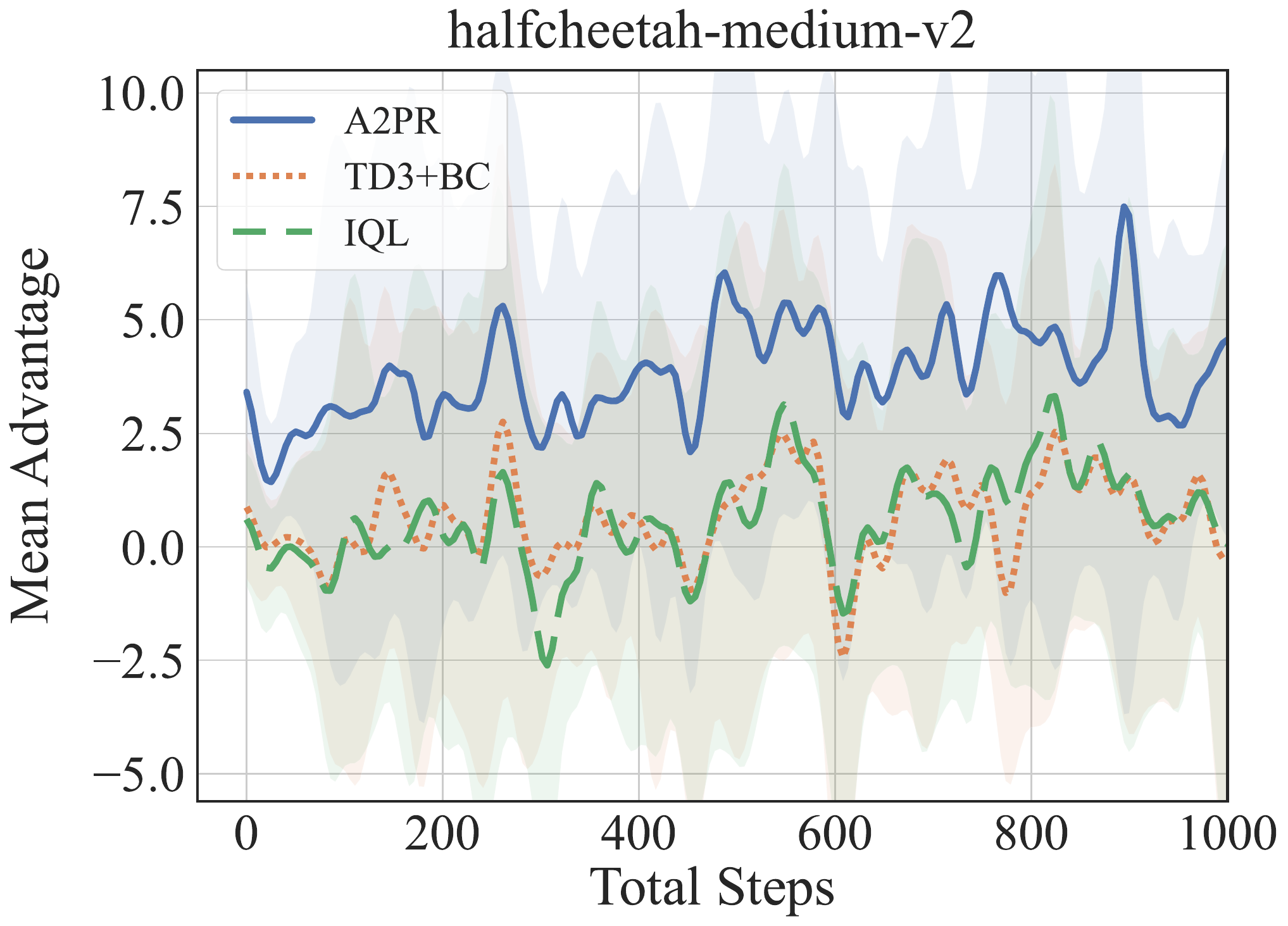}
		\end{minipage}%
	}%
	\subfigure[]{
		\begin{minipage}[t]{0.5\linewidth}
			\centering
			\includegraphics[width=3.25in]{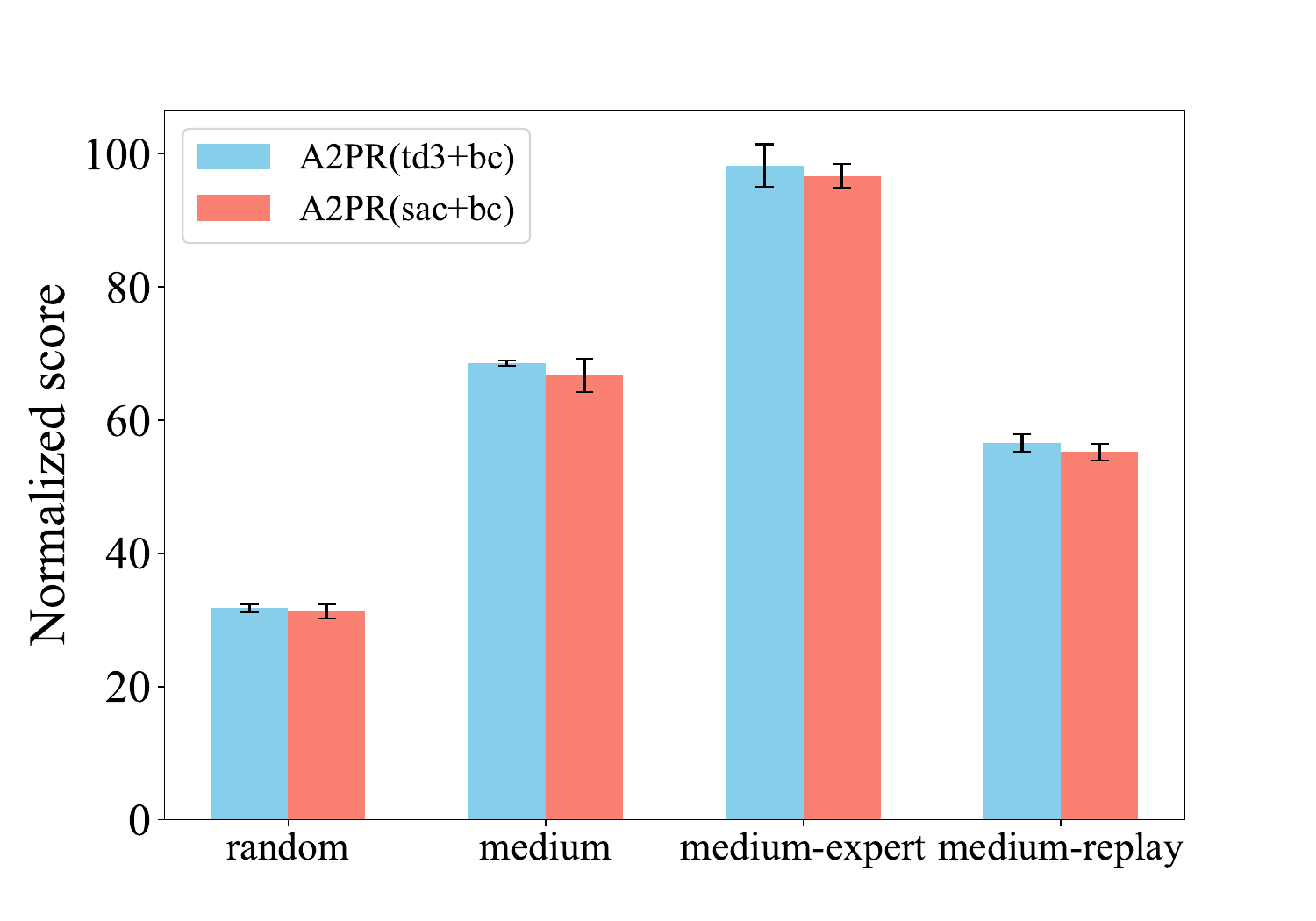}
		\end{minipage}%
	}%
    \caption{(a) The comparisons between the mean advantage curves of the actions of different methods on halfcheetah-medium-v2. (b) The performance comparison of A2PR on the TD3 variant and the SAC variant.
    }
    \label{fig:compar_adv}
\end{figure}

    We implement the A2PR algorithm on the Soft Actor-Critic (SAC)~\cite{haarnoja2018soft} framework, Which is compared to a version based on the TD3~\cite{fujimoto2018addressing} framework. Our experimental evaluation spans several datasets: halfcheetah-random-v2, halfcheetah-medium-v2, halfcheetah-medium-expert-v2, and halfcheetah-medium-replay-v2. The comparative analysis indicates that while the performance of the two approaches was broadly similar, the TD3 variant consistently outperformed the SAC variant, as shown in Figure \ref{fig:compar_adv}(b).



    
\subsection{Training time}
Managing time complexity presents a significant challenge in offline RL. We conduct experiments by running our methods and the baselines on the same dataset and machine for 1M steps. The re-training of TD3+BC, AWAC, IQL, and CQL was performed on the halfcheetah-medium-v2 dataset, utilizing implementations from \url{https://github.com/tinkoff-ai/CORL}~\cite{tarasov2022corl}. The re-training of PRDC was implemented using its official code from the original paper. The results in Table \ref{runtime-result} indicate that our method performs faster than other baselines on the halfcheetah-medium-v2 dataset, particularly when compared to CQL and PRDC, which require KD-tree for retrieval. Thanks to the efficiency of VAE's powerful generation, our method demonstrates notable speed.

\begin{table}[ht]
    \caption{The training time of the different methods}
    \vskip 0.10in
    \label{runtime-result}
	\centering
	\small
    \setlength{\tabcolsep}{4pt}
    \begin{tabular}{l|cccccccc}
    \hline
		\toprule
		\multicolumn{1}{c|}{\bf Methods}  & \multicolumn{1}{c}{\bf TD3+BC} & \multicolumn{1}{c}{\bf AWAC}  & \multicolumn{1}{c}{\bf IQL} &\multicolumn{1}{c}{\bf PRDC}&  \multicolumn{1}{c}{\bf CQL} &
        \multicolumn{1}{c}{\bf A2PR} \\ [2pt]
		\midrule
        Train time &2h18m  & 3h40m & 5h23m  & 6h49m  & 9h2m & 3h59m\\ [2pt]
		\bottomrule
  \hline
	\end{tabular}
\end{table}

\subsection{The generalization of A2PR on noisy datasets}
\begin{figure}[ht]
\begin{center}
    \centerline{\includegraphics[width=15cm]{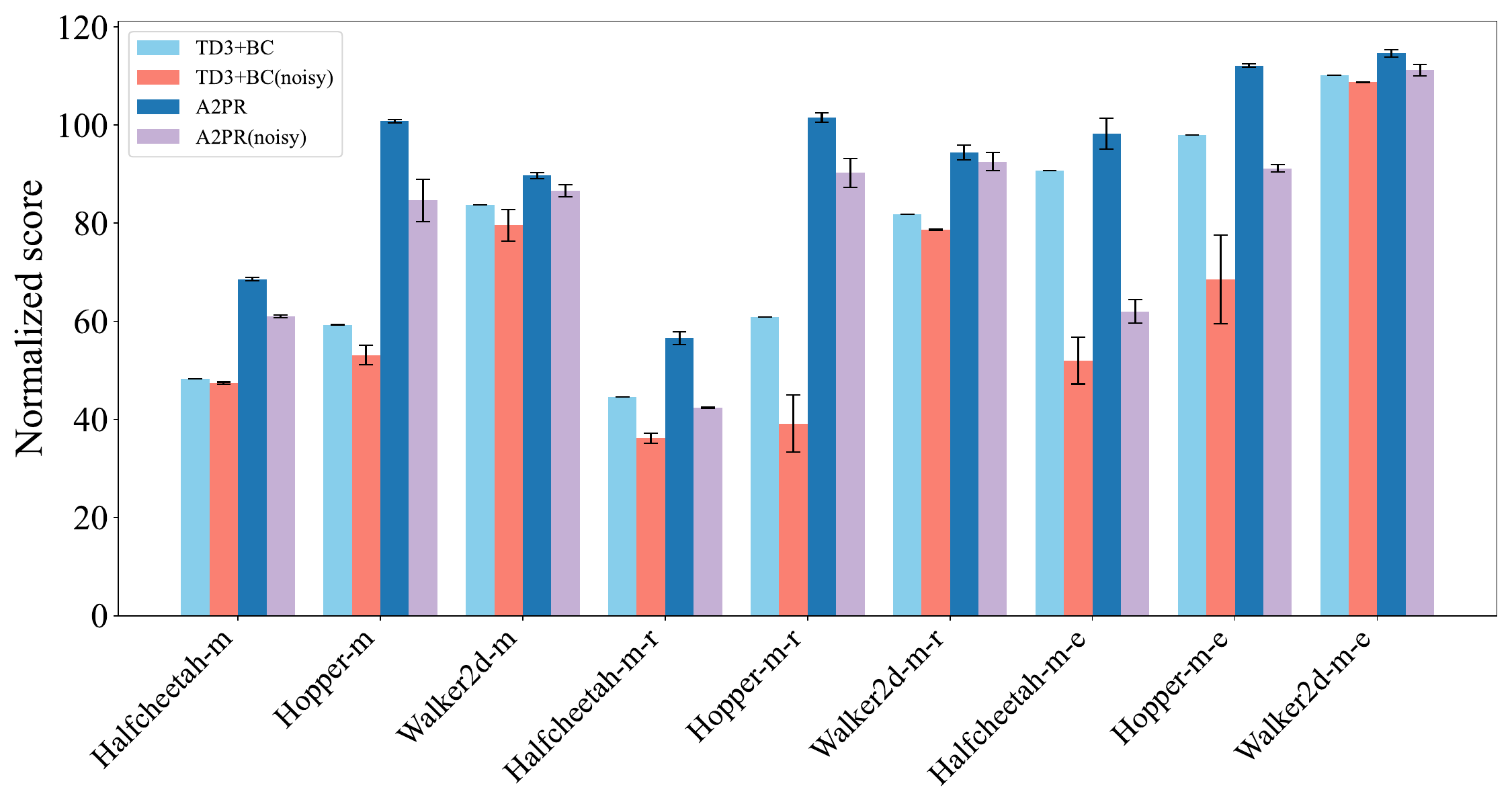}}
    \caption{The performance comparison of A2PR and TD3+BC on noisy datasets.
    }
    \label{fig:add_noisy}
    \end{center}
\end{figure}

We conducted experiments using the Multiple Target Maze and Mixed Random Policy Low-Quality datasets, illustrated in Figures \ref{fig:traj} and \ref{fig:mix_dataset}(a) in Section \ref{eva_addition}. These datasets differ significantly from those in the D4RL benchmark. Our results indicate that the A2PR algorithm outperforms established baselines, demonstrating noteworthy generalization capabilities.
To further assess the robustness and generalization performance of A2PR, we introduced Gaussian noise $\mathcal N(0,1)$  to the state inputs during the evaluation phase. We conduct experiments using the A2PR algorithm across three datasets: Hopper, HalfCheetah, and Walker2d (specifically -medium-v2, -medium-replay-v2,  -medium-expert-v2). For each dataset, the model was trained for 1 million steps across five different seeds. The results are shown in Figure \ref{fig:add_noisy}. This variant, A2PR(noisy), was compared against a similarly modified version of the TD3+BC algorithm, labeled TD3+BC(noisy), where noise was also added to the state inputs. The findings reveal that A2PR(noisy) not only consistently outperforms TD3+BC(noisy) but also maintains superior performance over the original TD3+BC across most tasks, despite a slight decrease in performance compared to its noise-free version. Notably, A2PR's performance on noisy states frequently surpasses that of TD3+BC on noise-free states, further underscoring A2PR's enhanced ability to generalize well under conditions of state perturbation.




\section{Implement details}
A2PR was implemented using PyTorch based on the TD3+BC implementation. The Elevating Positive Behavior Learning (EPBL) and Adaptive Advantage Policy Constraint (AAPC) components were implemented by ourselves. The hyperparameters of the algorithm are detailed in Table \ref{tab:hyperparameters} and Table \ref{hyperpara-value}.
To fully demonstrate the performance of A2PR and ensure fairness in comparison with the latest state-of-the-art research, the policy update equation introduces a new hyperparameter $w_2$ and retains the $\alpha$ values from the PRDC~\cite{ran2023policy} algorithm, which means retaining the same $\lambda$ values. 
    \begin{equation}\label{revised_eq}
        \mathcal{L}(\phi) = \mathbb E_{\substack{s,a\sim \mathcal{D},\\ \Bar{a}\in\{\tilde{a},\pi_\phi(s)\}}}\left[-\lambda Q_{\theta}(s,\pi_\phi(s)) + w_2
        (\pi_{\phi}(s)-\Bar{a})^2\right].
        \nonumber
    \end{equation}

\subsection{Hardware}
We use the following hardware:
\begin{enumerate}
    \item NVIDIA RTX 3090
    \item 12th Gen Intel(R) Core(TM) i7-12900K
\end{enumerate}
\subsection{Software} 
We use the following software versions:
\begin{enumerate}
    \item Python 3.9.19

    \item D4RL 1.1~\cite{fu2020d4rl}
    
    \item Mujoco 3.1.5~\cite{todorov2012mujoco}
    
    \item Gym 0.23.1~\cite{brockman2016openai}

    \item Mujoco-py 2.1.2.14

    \item Pytorch 1.13.1 + cu11.7~\cite{paszke2019pytorch}
\end{enumerate}
The v2 version of D4RL benchmark datasets is utilized in Gym locomotion and AntMaze tasks.

\begin{table}[t]

\caption{Hyperparameter Table}
\vskip 0.10in
\label{tab:hyperparameters}
\setlength{\tabcolsep}{21pt}
\centering
\begin{tabular}{c|c|c}
\hline
\toprule
& \textbf{Hyper-parameters} & \textbf{Value} \\
\midrule
\multirow{9}{*}{\textbf{TD3}} 
& Number of iterations & 1e6 \\
& Target update rate $\tau$ & 5e-3 \\
& Policy noise & 0.2 \\
& Policy noise clipping & (-0.5,0.5) \\
& Policy update frequency & 2 \\
& {Discount $\gamma$ for Mujoco} & 0.99, 0.995 \\
& {Discount $\gamma$ for Antmaze} & 0.995\\
& Actor learning rate &  3e-4 \\
& {Critic learning rate for Mujoco} & 3e-4 \\
& {Critic learning rate for Antmaze} & 1e-4\\
\hline
\multirow{11}{*}{\textbf{Network}} 
& Q-Critic hidden dim & 256\\
& Q-Critic hidden layers & 3  \\
& Q-Critic Activation function & ReLU \\
& V-Critic hidden dim & 256  \\
& V-Critic hidden layers & 3  \\
& V-Critic Activation function & ReLU \\
& Actor hidden dim & 256  \\
& Actor hidden layers & 2  \\
& Actor Activation function & ReLU \\
& Mini-batch size & 256 \\
& Optimizer & Adam~\cite{kingma2014adam} \\
\hline
\multirow{3}{*}{\textbf{A2PR}} & Normalized state & True \\
& {$\alpha$ for Mujoco} & 40.0, 2.5\\ 
& {$\alpha$ for Antmaze} & \{2.5, 7.5, 20.0\}\\ 
& {$\epsilon_A$ } & 0\\ 
		\bottomrule
\hline
\end{tabular}

\end{table}

\begin{table}[ht]
    \caption{Hyperparameter values for different tasks}
    \vskip 0.10in
    \label{hyperpara-value}
	\centering
    \setlength{\tabcolsep}{21pt}
	\begin{tabular}{l|cccccc}
    \hline
		\toprule
		\multicolumn{1}{c|}{\bf Task name}  & \multicolumn{1}{c}{ $w_1$} & \multicolumn{1}{c}{ $w_2$}  & \multicolumn{1}{c}{$\gamma$} &\multicolumn{1}{c}{$\alpha$} \\ [2pt]
		\midrule
        Halfcheetah-random-v2 & 1.0 & 1.0 & 0.99 & 40.0 \\ [2pt]
        Halfcheetah-medium-v2 & 1.0 & 1.0 & 0.99 & 40.0 \\ [2pt]
        Halfcheetah-medium-replay-v2 & 1.5 & 0.8 & 0.995 & 40.0 \\ [2pt]
        Halfcheetah-medium-expert-v2 & 1.0 & 15.0 & 0.99 & 40.0 \\ [2pt]
        Hopper-random-v2 & 1.5 & 1.5 & 0.995 & 2.5 \\ [2pt]
        Hopper-medium-v2 & 1.0 & 0.4 & 0.995 & 2.5 \\ [2pt]
        Hopper-medium-replay-v2 & 1.5 & 0.5 & 0.99 & 2.5 \\ [2pt]
        Hopper-medium-expert-v2 & 1.0 & 4.0 & 0.99 & 2.5 \\ [2pt]
        Walker2d-random-v2 & 1.0 & 1.0 & 0.995 & 2.5 \\ [2pt]
        Walker2d-medium-v2 & 1.5 & 1.0 & 0.99 & 2.5 \\ [2pt]
        Walker2d-medium-replay-v2 & 1.5 & 1.5 & 0.99 & 2.5 \\ [2pt]
        Walker2d-medium-expert-v2 & 1.0 & 0.8 & 0.99 & 2.5 \\ [2pt]
		\bottomrule
    \hline
	\end{tabular}
\end{table}


\end{document}